\documentclass{article}

\usepackage[final]{neurips_2025}

\usepackage[utf8]{inputenc} 
\usepackage[T1]{fontenc}    
\usepackage[hidelinks, urlcolor=black]{hyperref}      
\usepackage{url}            
\usepackage{booktabs}       
\usepackage{amsfonts}       
\usepackage{nicefrac}       
\usepackage{microtype}      
\usepackage{xcolor}         
\usepackage{bbding}
\usepackage{graphicx}
\usepackage{amsmath}
\usepackage{subcaption}
\usepackage{multirow}    
\usepackage{booktabs}    
\usepackage{amsmath}     
\usepackage{textcomp}    
\usepackage{enumitem}
\usepackage[autostyle]{csquotes} 

\makeatletter
\providecommand{\@trackname}{}  
\makeatother

\definecolor{red}{rgb}{1,0,0} 

\title{RS3DBench: A Comprehensive Benchmark for 3D Spatial Perception in Remote Sensing}

\author{
Jiayu Wang\,$^{1,*}$, Ruizhi Wang\,$^{2,*}$, Jie Song\,$^{1}$, Haofei Zhang\,$^{1}$, Mingli Song\,$^{1}$, Zunlei Feng\,$^{1}$, Li Sun\,$^{1}$\\
$^1$ College of Computer Science and Technology, Zhejiang University, Hangzhou, China\\
$^2$ Software College of Zhejiang University, Ningbo, China\\
$^3$ Hangzhou City University, Hangzhou, China\\
\texttt{22451097@zju.edu.cn}, \texttt{ruizhiwang@zju.edu.cn}, \texttt{sjie@zju.edu.cn}, \texttt{haofeizhang@zju.edu.cn},\\
\texttt{brooksong@zju.edu.cn}, \texttt{zunleifeng@zju.edu.cn}, \texttt{lsun@zju.edu.cn}
}

\begin{document}

\maketitle

\let\thefootnote\relax\footnotetext{* These authors contributed equally to this work.}

\begin{abstract}
  In this paper, we introduce a novel benchmark designed to propel the advancement of general-purpose, large-scale 3D vision models for remote sensing imagery. While several datasets have been proposed within the realm of remote sensing, many existing collections either lack comprehensive depth information or fail to establish precise alignment between depth data and remote sensing images. To address this deficiency, we present a visual \textbf{Bench}mark for \textbf{3D} understanding of \textbf{R}emotely \textbf{S}ensed images, dubbed \textbf{RS3DBench}. This dataset encompasses 54,951 pairs of remote sensing images and pixel-level aligned depth maps, accompanied by corresponding textual descriptions, spanning a broad array of geographical contexts. It serves as a tool for training and assessing 3D visual perception models within remote sensing image spatial understanding tasks. Furthermore, we introduce a remotely sensed depth estimation model derived from stable diffusion, harnessing its multimodal fusion capabilities, thereby delivering state-of-the-art performance on our dataset. Our endeavor seeks to make a profound contribution to the evolution of 3D visual perception models and the advancement of geographic artificial intelligence within the remote sensing domain. The dataset, models and code will be accessed on the \url{https://rs3dbench.github.io}.
  
\end{abstract}

\section{Introduction}\label{sec:intro}

Remote sensing systems~\cite{47} have emerged as pivotal tools for large-scale geospatial intelligence, leveraging multi-spectral satellite imagery to decode the Earth's surface with centimeter-to-meter scale resolution. Beyond mere observation, these systems enable predictive analytics across strategic domains: automated land use mapping~\cite{48} powers sustainable resource management; urban 3D reconstruction~\cite{49} drives smart city development; while crop phenotyping via hyperspectral imaging~\cite{50} and disaster damage assessment~\cite{51} demonstrate their societal impact.
\begin{figure}[htbp]
    \centering
    \includegraphics[width=1\textwidth]{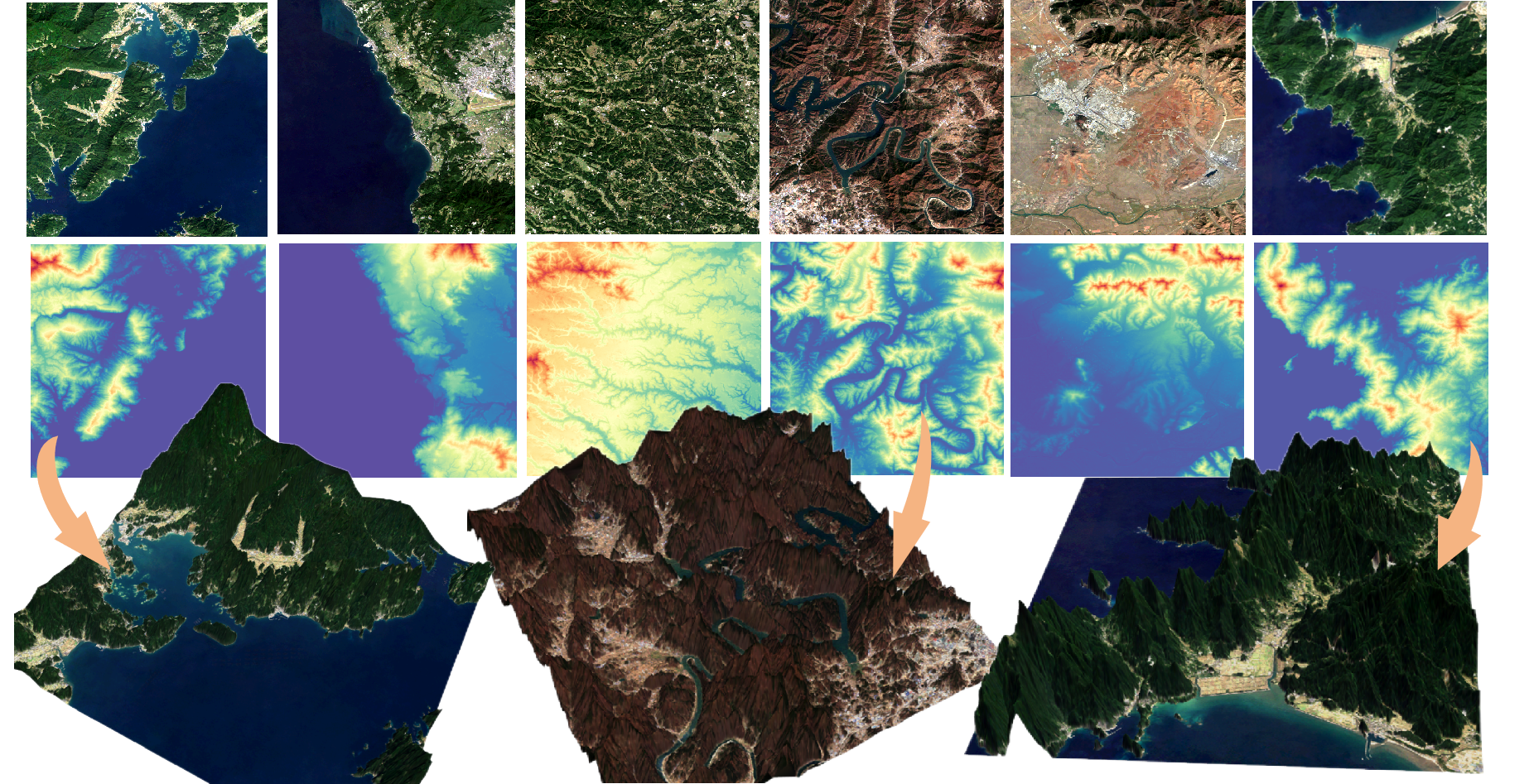}
    \caption{We proposed RS3DBench: A comprehensive remote sensing benchmark with 54,951 pairs of RGB-DEM. This shows the pixel-aligned RGB-DEM pairs included of RS3DBench with corresponding generated 3D terrain reconstructions.}
\end{figure}
Beyond these applications, remote sensing data holds transformative potential for visual-language navigation (VLN), with indoor benchmarks such as R2R~\cite{53} and VLN-CE~\cite{54} demonstrating considerable success. However, expanding agent interaction environments from indoor to outdoor scenarios necessitates the comprehension of large-scale terrain morphology and complex multi-altitude spatial relationships, encountering a critical bottleneck: the absence of spatially anchored datasets capable of bridging aerial perspective reasoning with terrain-level semantics.

Height information, as the cornerstone of 3D geospatial reasoning, remains constrained by sensor physics: Airborne LiDAR delivers centimeter-level precision at operational costs exceeding hundreds of dollars/km², whereas InSAR and PRISM stereo imaging necessitate meticulous flight planning and multi-sensor calibration~\cite{52}. Though accurate, these methods prove economically and temporally prohibitive for global deployment, further hampered by environmental constraints (\textit{e.g.}, cloud obstruction, vegetation interference). In recent years, 3D visual perception models grounded in deep learning have precipitated a paradigmatic shift in the analysis of natural scene structures~\cite{1,2,3,4,5,6,7,8}, achieved through synergistic optimization of multimodal data fusion (encompassing RGB-D inputs, LiDAR point clouds, and multi-view imagery) and geometric prior encoding~\cite{21,22,23,24,25,26}. While demonstrating exceptional scene reconstruction capabilities with close-range RGB-D data (exemplified by NYU Depth V2~\cite{9} and ScanNet~\cite{10}), the direct transposition of these 3D perception models to the remote sensing domain encounters formidable domain adaptation challenges. Ordinary scenes and remote sensing scenes are very different. Remote sensing images often contain very small objects (sometimes only 10 pixels) and require complex spatial reasoning from a top-down view. These obstacles originate from two fundamental issues:

(\textbf{i}) \textbf{Data Distribution Disparity.}
The domain discrepancy between proximal sensing modalities (\textit{e.g.}, indoor RGB-D sensors, vehicular LiDAR) and satellite remote sensing systems originates from fundamental differences in imaging physics that manifest most acutely in scale-space representations~\cite{35,36}, particularly in scale representation. Critical ground objects (\textit{e.g.}, vehicles, minor infrastructure) span a mere 10$\sim$20 pixels in sub-meter resolution images, rendering conventional convolutional kernel-based feature extractors inadequate for capturing salient geometric features~\cite{27}.

(\textbf{ii}) \textbf{Absence of Depth Information}:
Current remote sensing datasets exhibit significant limitations in 3D perception tasks due to: (1) ‌lack of depth annotations, and (2) misalignment of depth and RGB data. For the former, predominant remote sensing datasets furnish only 2D bounding boxes and categorical labels~\cite{11,15}, devoid of elevation or surface geometry descriptors. This constraint compels models to resort to weakly supervised approaches (\textit{e.g.}~\cite{28}, inferring height from shadow length). For the latter, even the limited datasets containing depth information frequently lack corresponding RGB imagery with pixel-level registration, further exacerbating the challenge of accurate 3D reconstruction.

The construction of an efficacious remote sensing 3D perception model necessitates a comprehensive, high-fidelity dataset as its foundation. Addressing these limitations, our research introduces RS3DBench, a new universal benchmark that includes 54,951 pairs of pixel-aligned remote sensing images, depth information, and text descriptions, covering multiple countries across four continents, as well as six major terrain types for 3D visual understanding of remote sensing images. This dataset facilitates the training and rigorous evaluation of 3D vision models across diverse remote sensing interpretation tasks. The key contributions of our work are summarized as follows:
\begin{itemize}[leftmargin=*]
\item[$\bullet$] We propose an innovative semi-automated 3D vision data acquisition pipeline comprising four critical stages: RGB and depth data capture, depth alignment, GLM-v4 inference, and human validation. This framework enables efficient compilation of extensive, high-quality datasets enriched with depth information.
\item[$\bullet$] Leveraging this semi-automated pipeline, we have assembled 54,951 paired remote sensing images with pixel-aligned depth maps, accompanied by comprehensive textual descriptions spanning diverse geographical contexts. This corpus enables exhaustive evaluation of various 3D visual perception models' capabilities.
\item[$\bullet$]We proposed a novel multimodal fusion method that integrates geographic semantic information from remote sensing images into the diffusion process of a pretrained stable diffusion model for depth estimation in the form of text, achieving state-of-the-art performance on our dataset.
\item[$\bullet$]We built a rigorous benchmarking framework based on the RS3DBench dataset, developed a comprehensive evaluation protocol, and evaluated the performance of several state-of-the-art depth estimation models.
\item[$\bullet$]Our study establishes that geosemantic text embeddings can effectively enhance remote sensing depth estimation through diffusion frameworks, with rigorous validation conducted on the RS3DBench benchmark.
\end{itemize}

\section{Related Work}
\noindent\textbf{Remote sensing dataset.} Existing remote sensing datasets focus on image-text pairs and image annotation frameworks~\cite{34}.Terrain analysis~\cite{29} and large-scale 3D reconstruction~\cite{30}, however, demand datasets with pixel-level spatial alignment to RGB imagery, multi-resolution, global geographic diversity, and fine-grained terrain representation, thereby accounting for the acute scarcity of high-quality geospatial dataset. Notably, even state-of-the-art 3D perception paradigms struggle to bridge this gap, as their reliance on localized close-range data (\textit{e.g.}, NYU Depth V2~\cite{9}) conflicts with sub-10-pixel object localization imperatives of remote sening analysis.
\begin{table}[ht]
  \caption{Compared with multimodal remote sensing datasets, RS3DBench is the first remote sensing dataset with aligned spatial information, and Align represents the alignment of image-to-depth information. Multi-Reg represents coverage across multiple global regions.}
  \label{tab:dataset_comparison}
  \centering
  \begin{tabular}{l|c|c|c|c|c|c|c}
    \toprule
    \textbf{Dataset}       & \textbf{Image} & \textbf{Text} & \textbf{Depth} & \textbf{Align} & \textbf{Multi-Res} & \textbf{Multi-Reg} & \textbf{Scene} \\
    \midrule
    UCM~\cite{11}                & $\checkmark$ & \XSolidBrush &  \XSolidBrush      &  \XSolidBrush     &  \XSolidBrush      &  \XSolidBrush           & Mixed      \\
    RSICD~\cite{12}             & $\checkmark$ & $\checkmark$ &  \XSolidBrush      &  \XSolidBrush     & $\checkmark$      & $\checkmark$           & Mixed      \\
    NWPU-cap~\cite{13}            & $\checkmark$ & $\checkmark$ &  \XSolidBrush      &  \XSolidBrush     & $\checkmark$      & $\checkmark$           & Mixed      \\
    RSSM~\cite{14}                & $\checkmark$ & $\checkmark$ &  \XSolidBrush      &  \XSolidBrush     & $\checkmark$      & $\checkmark$           & Mixed      \\
    WHU-OPT-SAR~\cite{15}         & $\checkmark$ &  \XSolidBrush &  \XSolidBrush      &  \XSolidBrush     & $\checkmark$      & $\checkmark$           & Mixed      \\
    swissALTI3D~\cite{16}              &  \XSolidBrush &  \XSolidBrush & $\checkmark$      &  \XSolidBrush     &  \XSolidBrush      & $\checkmark$          & Nature     \\
    AW3D30~\cite{17}             &  \XSolidBrush &  \XSolidBrush & $\checkmark$      &  \XSolidBrush     &  \XSolidBrush      & $\checkmark$          & Nature     \\
    UrbanScene3D~\cite{19}        & $\checkmark$ &  \XSolidBrush & $\checkmark$      & \XSolidBrush    &  \XSolidBrush      &  \XSolidBrush           & Urban      \\
    MatrixCity~\cite{18}         & $\checkmark$ &  \XSolidBrush & $\checkmark$      & \XSolidBrush     &  \XSolidBrush      &  \XSolidBrush           & Urban      \\
    KITTI Depth~\cite{20}         & $\checkmark$ &  \XSolidBrush & $\checkmark$      &  $\checkmark$     &  \XSolidBrush      &  \XSolidBrush       & Driving    \\
    NYU Depth V2~\cite{9}        & $\checkmark$ &  \XSolidBrush & $\checkmark$      &  $\checkmark$     &  \XSolidBrush      &  \XSolidBrush          & Indoor     \\
    \midrule
    \textbf{Ours}      & $\checkmark$ & $\checkmark$ & $\checkmark$      & $\checkmark$     & $\checkmark$      & $\checkmark$ & \textbf{Nature} \\
    \bottomrule
  \end{tabular}
\end{table}
As shown in Table~\ref{tab:dataset_comparison}, existing remote sensing images such as the datasets RSICD~\cite{12} and NWPU-CAP~\cite{13}  also try to add detailed text descriptions to remote sensing images, but the scale is small and lacks sufficient diversity. Although RS5M~\cite{14}  constructs a large-scale remote sensing image-text matching dataset, it still lacks spatial information, while the WHU-OPT-SAR~\cite{15} dataset contains optical (OPT) and synthetic aperture radar (SAR) images to provide complementary information, but does not align the depth information and remote sensing images, and the dataset has a single resolution. Public DEM datasets such as AW3D30~\cite{17}  and swissALTI3D~\cite{16}  only provide elevation data and do not have pixel-level registration with remote sensing images. UrbanScene3D~\cite{19}  and MatrixCity~\cite{18} align RGB images with spatial information to generate 3D models of urban scenes, but they only target the city scenes, the data for natural scenes. NYU Depth V2~\cite{9} and KITTI Depth~\cite{20} demonstrate the importance of RGB-D data for 3D understanding, but the paradigm is difficult to transfer directly from ordinary scenes to remote sensing scenes and multi-resolution characteristics of satellite images are fundamentally different from the local perspectives of natural scenes.

In this paper, we propose the first remote sensing dataset with spatial information and remote sensing image alignment, with multiple resolutions, covering a variety of terrain and landforms, and corresponding text descriptions, and our core contribution is to pioneer the pixel-level spatiotemporal alignment of spatial information and remote sensing data. As the differentiated feature of this dataset, spatial information solves the pain point of the lack of spatial information in traditional datasets, provides a key geometric prior for tasks such as 3D terrain modeling and dynamic environment simulation, and lays a data cornerstone for 3D geospatial intelligence applications such as digital cities and geological disaster early warning.

\noindent\textbf{Remote sensing depth estimation.}
Digital Elevation Model (DEM) serves as the core data source to support hydrogeographic analysis, 3D spatial planning and geographic information systems. Traditional methods rely on technologies such as LiDAR, Synthetic Aperture Radar Interferometry (InSAR), and Stereo Imaging Sensor (PRISM) to obtain high-precision topographic data, but they are subject to complex acquisition processes, including engineering challenges such as flight parameter planning, equipment calibration and maintenance, and multi-sensor collaboration~\cite{43,44,45,46}, resulting in high time and economic costs. In this context, the end-to-end depth estimation of remote sensing images significantly improves the efficiency of DEM generation.

In the field of deep learning, since Eigen et al.~\cite{37} pioneered multi-scale convolutional networks, depth estimation models have continued to evolve: the introduction of structured priors such as planar guidance maps~\cite{38}, piecewise planarity priors~\cite{40} and neural conditional random fields~\cite{39} has strengthened geometric consistency, variational autoencoders~\cite{41} and adaptive discretization strategies (such as AdaBins~\cite{8} and BinsFormer~\cite{42}) have promoted refined modeling, and visual transformers have enhanced global context perception through self-attention mechanisms. The diffusion model~\cite{31} uses generative priors to achieve geometrically coherent depth synthesis. However, although mainstream models such as Marigold~\cite{1} and HDN~\cite{5} perform well in conventional scenes, they are difficult to cope with the extreme scale changes caused by the unique 10-meter pixel resolution of remote sensing images (for example, only 10-20 pixels are left to represent the landform features such as ridges and water systems) and complex occlusion. Due to the scarcity of urban scene data~\cite{1,30}and the lack of natural terrain modeling, it is urgent to establish cross-scenario evaluation benchmarks. In this study, we break through the limitations of traditional indoor/urban structured scenarios, extend depth estimation to wide-area remote sensing perspectives and natural terrain analysis, and construct systematic evaluation to reveal the potential and limitations of existing models in complex landform modeling.
\section{Proposed RS3DBench Dataset}

\label{headings}
\begin{figure}[htbp]
    \centering
    \includegraphics[width=1\textwidth]{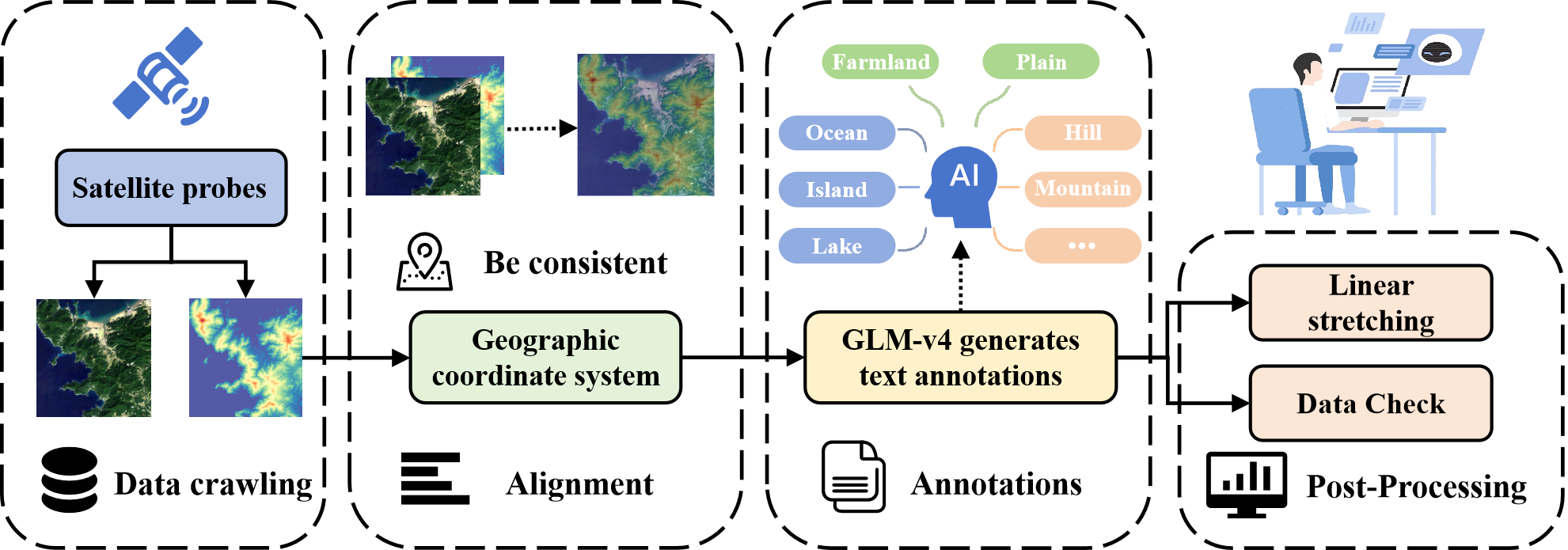}
    \caption{The pipeline of data collection of RS3DBench:  
(1) \textbf{Data Crawling}  
(2) \textbf{Alignment}
(3) \textbf{Annotations} 
(4) \textbf{Post-Processing}}
    \label{fig:pipeline}
\end{figure}

We introduce the RS3DBench dataset, a remote sensing dataset comprising pixel-aligned RGB-DEM pairs at four distinct resolutions (30m, 10m, 2m, and 0.5m), offering comprehensive global coverage across a variety of terrain types. This dataset is designed to facilitate depth estimation tasks within the realm of remote sensing and is available under the CC BY-SA 4.0 license. Figure~\ref{fig:pipeline} illustrates the pipeline for creating RS3DBench, with a thorough explanation of each step provided in the subsequent sections. We then present the statistics of dataset to offer a clearer understanding of RS3DBench.

\subsection{Pipeline of Constructing the RS3DBench Dataset}

\noindent\textbf{Data crawling.} 
The RS3DBench dataset amalgamates multi-source DEM data: ALOS AW3D30 (30m resolution for Japan, Korea, Southeast Asia, and the Mediterranean)~~\cite{17}, SwissALTI3D LiDAR fusion data (with a resolution ranging from 0.5 to 2m)~~\cite{16}, and Sentinel-2-derived Australian DEM (5m)~~\cite{33}. Standardized according to the UTM/WGS84 coordinate systems, this geospatial repository encompasses coastal lowlands, highlands, mountainous regions, and plains, thus enabling large-scale terrain analysis and precise surface modeling. 
For more details, please refer to Appendix~\ref{app:Depth}.

\noindent\textbf{Alignment.} 
We have solved the key issue of the lack of RGB data in the public digital elevation model (DEM) dataset (ALOS/swissALTI3D)~\cite{17,16}. We obtained the geographic coordinate locations and coordinate system IDs of the DEM data, and converted them to location coordinates in the 'EPSG:4326' or 'EPSG:2056' coordinate systems, in order to retrieve the corresponding RGB data provided by Sentinel 1, Sentinel 2, swissALTI3D, and Google Earth. The RGB images will undergo cloud and haze removal, atmospheric correction, radiometric correction, and normalization to create rigorously aligned DEM and RGB data. Considering that these data do not strictly align with RGB images in terms of temporal and spatial dimensions, and that some images are turbid and blurry, about 20\% to 30\% of the misaligned images were manually screened and filtered. Through experiments, we identified outliers generated due to the limitations of satellite sensors that created misleading models, and processed them through interpolation to ensure the accuracy of the data. For more details, please refer to Appendix~\ref{app:text}.

\noindent\textbf{Annotations.} \label{sec:text}
We use the large language model GLM-v4~\cite{32} to add text annotations to remote sensing images and introduce a constraint mechanism in the process to relate it to terrain classification and landform description, guiding the model to generate more accurate descriptions. By embedding domain knowledge and implementing filtering strategies, we significantly improve the relevance between the generated text and remote sensing images. To enhance the quality of text generation, we iterated on the input prompts for GLM-v4 multiple times, ultimately establishing a specialized prompt framework for remote sensing. To ensure the effectiveness and reliability of text descriptions, we have developed a comprehensive manual and automated review mechanism that monitors potential issues arising from timeouts or network errors with the large model API in real time. Additionally, unreasonable annotations are re-annotated manually, and we analyze and optimize the causes of incorrect annotations during the annotation process to facilitate timely improvements in design and reprocess erroneous samples. Successfully, semantic accurate, context-relevant, and high-quality text descriptions have been generated for all remote sensing image pairs. The generated descriptions adequately reflect the spatial features and details of the characteristics in the images, providing a reliable guarantee for downstream tasks. For more details on text annotations, please refer to the Appendix~\ref{app:text}.

\noindent\textbf{Post-Processing.}
We resized sample images to 512 × 512 pixels. To address brightness issues caused by atmospheric interference or cloud occlusion in the original Sentinel-2 imagery, we designed a three-stage image enhancement process. (1) Normalize the three RGB channels separately to the range of 0-255. (2) Based on the set percentile range, the pixel values in the images were restricted to that range, replacing anomalous extreme values with extreme values within the range. (3) The pixel values of the images were linearly remapped from the current brightness range back to 0-255 to reduce color distortion. This three-stage optimization improves image quality. The implementation details are provided in the Appendix~\ref{app:linear}.

For each pair of RGB-DEM with text annotations, we will conduct a manual inspection: (1) the quality of RGB and DEM; (2) whether the RGB image is fully aligned with the DEM; (3) check the rationality of the text annotations.

\subsection{Dataset statistics and analysis}

\begin{figure}[htbp]
  \centering 
  
  \begin{subfigure}[t]{0.32\textwidth}
    \includegraphics[width=\textwidth, height=5cm, keepaspectratio]{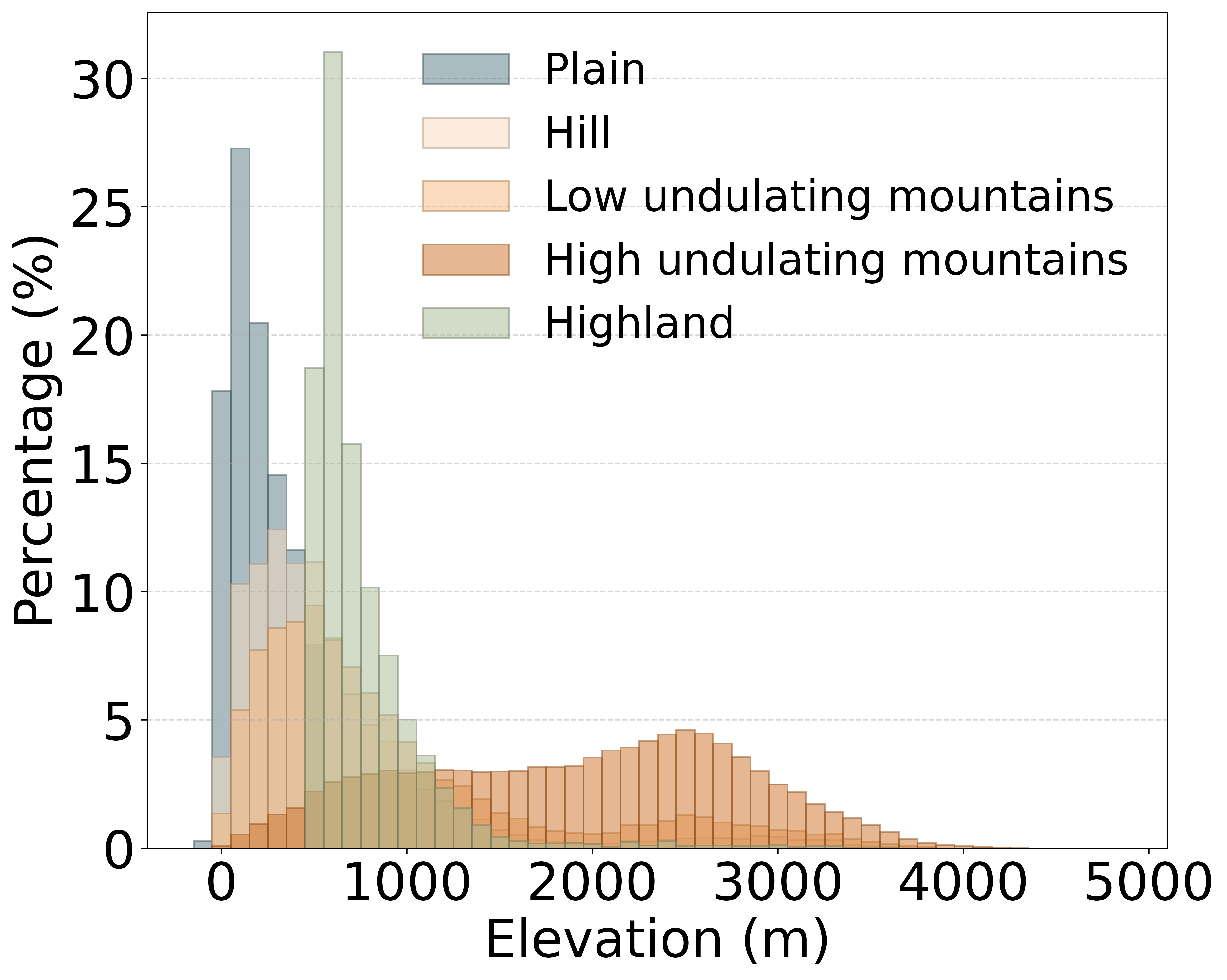}
    \caption{Elevation distribution by terrain}
    \label{fig:sub1}
  \end{subfigure}
  \hfill
  \begin{subfigure}[t]{0.32\textwidth}
    \includegraphics[width=\textwidth, height=5cm, keepaspectratio]{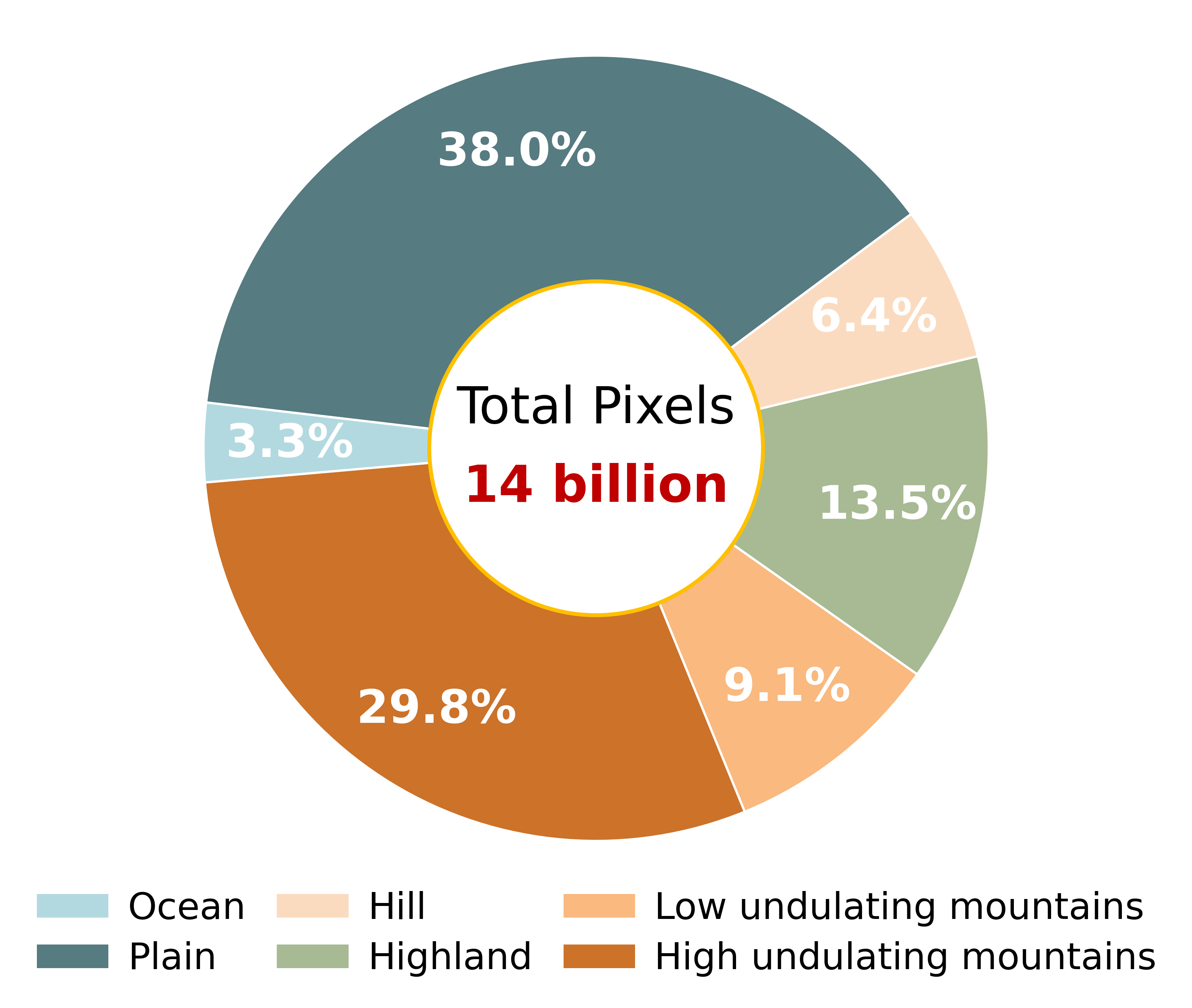}
    \caption{Terrain type distribution}
    \label{fig:sub2}
  \end{subfigure}
  \hfill
  \begin{subfigure}[t]{0.32\textwidth}
    \includegraphics[width=\textwidth, height=5cm, keepaspectratio]{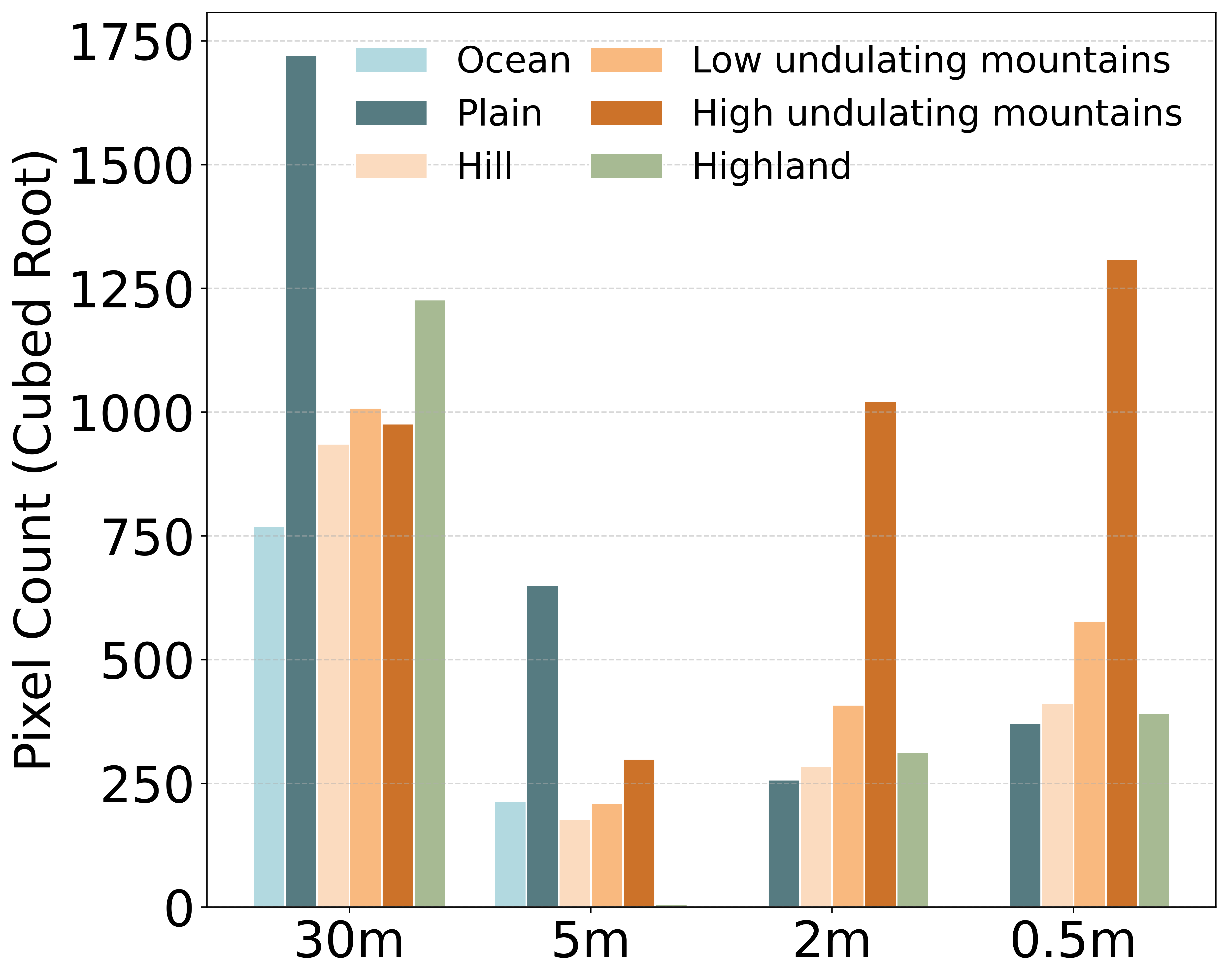}
    \caption{Depth value statistics}
    \label{fig:sub3}
  \end{subfigure}
  \caption{The statistical distribution of dataset resolution, terrain, depth data}
  \label{fig:dataset_stats}
\end{figure}
Figure~\ref{fig:dataset_stats} (a) shows the distribution percentage of different terrains in the dataset at various altitudes, excluding the ocean. The main altitude of the ocean is 0, and the altitude range is distributed between -149 to 4813 meters, which is of great significance for large-scale remote sensing terrain modeling, effectively compensating for the shortcomings of existing depth estimation datasets in terms of large elevation changes and multiple terrain scenarios. Figure~\ref{fig:dataset_stats} (b) presents the overall proportion of the six major terrain categories: ocean 3.3\%, low undulating mountains 9.1\%, hills 6.4\%, plains 38.0\%, Highland 13.5\%, and high undulating mountains 29.8\%, providing crucial support for model training in complex geographical scenarios. Figure~\ref{fig:dataset_stats} (c) depicts the distribution of terrain counts at four resolutions (30 m, 5 m, 2 m, 0.5 m) with the cube root of the number of pixels as the vertical axis. The results indicate that low resolution (30 m, 5 m) datasets have a higher proportion of plains and Highland, while high-resolution (2 m, 0.5 m) primarily consists of mountainous types. For more details, please refer to Appendix~\ref{app:Depth}.

\section{Marigold-RS}\label{sec:Method}
\subsection{Generative Formulation}\label{sec:GF}
We pose the depth estimation of remote sensing images as a denoising diffusion process for text and image conditions~\cite{31} , extending the original framework of Marigold~\cite{1} . The goal is to model the conditional distribution \( D(\mathbf{d} \mid \mathbf{x}, E) \), where \(\mathbf{d} \in \mathbb{R}^{W \times H}\) is the depth map, \(\mathbf{x} \in \mathbb{R}^{W \times H \times 3}\) is the input RGB image, and \(E\) represents the text annotations (\textit{e.g.}, scene categories).  

Similar to Marigold, the forward process adds Gaussian noise to the depth map \(\mathbf{d}_0\), and the reverse process learns to denoise \(\mathbf{d}_t\) at timestep \(t\). Formally, given an RGB condition image \(\mathbf{x} \in \mathbb{R}^{W \times H \times 3}\), the forward process corrupts the ground truth depth \(\mathbf{d}_0\) over \(T\) steps by iteratively applying noise according to a predefined variance schedule \(\{\beta_1, ..., \beta_T\}\):
\begin{equation}
\mathbf{d}_t = \sqrt{\bar{\alpha}_t} \mathbf{d}_0 + \sqrt{1 - \bar{\alpha}_t} \epsilon, \quad \epsilon \sim \mathcal{N}(0, I),
\end{equation}
where \(\bar{\alpha}_t := \prod_{s=1}^t (1-\beta_s)\) controls the noise level at step \(t\). During the reverse process, a conditional denoising model \(\epsilon_\theta(\cdot)\) parameterized by \(\theta\) is trained to remove noise \(\mathbf{d}\) to obtain \(\mathbf{d}_{t-1}\):
\begin{equation}\mathbf{d}_{t-1} = \frac{1}{\sqrt{1-\beta_t}} \left( \mathbf{d}_t - \frac{\beta_t}{\sqrt{1-\bar{\alpha}_t}} \epsilon_\theta( \mathbf{z}_t^{(\mathbf{d})}, \mathbf{z}^{(\mathbf{x})}, \tau_\theta(E), t) \right) + \sigma_t \mathbf{z}, \quad \mathbf{z} \sim \mathcal{N}(0, I),
\end{equation}
\(\sigma_t\) governing the stochasticity. However, our formulation introduces textual conditioning alongside image conditioning. During training, the denoising U-Net \(\epsilon_\theta\) is optimized to predict noise not only from the noisy depth latent \(\mathbf{z}_t^{(\mathbf{d})}\) and image latent \(\mathbf{z}^{(\mathbf{x})}\), but also from the text embedding \(\tau_\theta(E)\). The objective function becomes~\cite{32}:  
\begin{equation}
\mathcal{L} = \mathbb{E}_{\mathbf{d}_0, \epsilon, t} \left\| \epsilon - \epsilon_\theta \left( \mathbf{z}_t^{(\mathbf{d})}, \mathbf{z}^{(\mathbf{x})}, \tau_\theta(E), t \right) \right\|_2^2.
\end{equation}
This extension allows the model to resolve ambiguity in depth estimation by utilizing semantic priors in text descriptions (\textit{e.g.}, \enquote{mountains} or \enquote{oceans}), especially for remote sensing scenarios that are complex and have distinct class characteristics. For training details, see Appendix ~\ref{app:implementation}.

In the inference stage, starting from a normally-distributed variable \(\mathbf{d}_{T}\), the learned denoiser is iteratively applied to remove the noise step by step until \(\mathbf{d}_{0}\) is reconstructed.

\subsection{Depth estimation framework with text guidance}\label{sec:Depth_frame}

Our architecture is built on the Marigold model, integrates text conditioning into depth estimation through three key components:

\textbf{Text preprocessing and encoding.} We use the pre-trained CLIP to encode \(E\) into \(\tau_\theta(E)\):
\begin{equation}
\tau_\theta(E)\ = \text{TextEncoder}(E) \in \mathbb{R}^{B \times L \times D},
\end{equation}
where \(B\), \(L\) and \(D\) correspond to batch size, sequence length, and embedding dimension respectively.

\textbf{Cross-Modal Fusion.} We leverage the pretrained VAE to encode both RGB images
\(\mathbf{x}\) and single-channel depth maps \(\mathbf{d}\) into a shared latent space as \(\mathbf{z}^{(\mathbf{x})}\) and \(\mathbf{z}^{(\mathbf{d})}\). For depth encoding, the 1-channel map is triplicated to 3-channel (RGB) input dimensions, enabling direct compatibility with the pretrained encoder. We integrate text embeddings in the following ways:
\begin{equation}
  \mathbf{z}_t = \text{CrossAttention}\Big( 
    \text{Concat}(\mathbf{z}^{(\mathbf{x})}, \mathbf{z}_t^{(\mathbf{d})}), 
    \tau_\theta(E)\
  \Big).
\end{equation}
\textbf{Noise residual prediction.}
In the final layers of the U-Net, after integrating the temporal embedding \( \tau(t) \) (derived from timestep \( t \)) and the multi-scale features from cross-attention, the model predicts the noise residual through a series of convolutional, normalization, and activation layers:
\begin{equation}
\hat{\epsilon} = \epsilon_\theta\left( \mathbf{z}_t, t \right) \in \mathbb{R}^{B \times C \times H \times W},
\end{equation}
where \( \mathbf{z}_t \) represents the fused latent features combining the noisy depth \(\mathbf{z}_t^{(\mathbf{d})}\), image condition \(\mathbf{z}^{(\mathbf{x})}\), and text embeddings \(\mathbf{z}^{(E)}\) through cross-modal fusion.
\section{Benchmarks}\label{sec:Exper}
In this section, to evaluate the quality of RS3DBench and its data support for remote sensing image depth estimation tasks, we conducted a comprehensive assessment from five aspects. We evaluated RS3DBench using state-of-the-art methods based on various backbones. We used three metrics to assess the results: \text{MAE}, $\delta$ threshold accuracy and \text{RMSE}, Please refer to Appendix~\ref{app:appendixB} for details.



\subsection{Cross-Domain Adaptability of Traditional Depth Models in Remote Sensing}
In this study, We focus on the task of depth estimation from remote sensing RGB images. In our experiments, we trained and evaluated six depth estimation models based on different backbones. We trained and evaluated six depth estimation models based on different backbones, as well as our method, on the 30m accuracy dataset of the RS3DBench dataset.
\begin{table}[ht]
  \caption{In order to verify the effectiveness of the dataset on the depth estimation model.}
  \label{tab:method_comparison}
  \centering
  \begin{tabular}{l|c|c|c|c|c|c|c}
    \toprule
    \textbf{Model}       & \textbf{Backbone} & \textbf{Params} & \textbf{MAE$\downarrow$} &\textbf{RMSE$\downarrow$} & \textbf{\boldmath$\delta$$\uparrow$} & \textbf{\boldmath$\delta^{2}$$\uparrow$} & \textbf{\boldmath$\delta^{3}$$\uparrow$} \\
    \midrule
    pix2pix~\cite{4}        &GAN &  207M  & 34.8 &   42.4 & 32.7   &   56.5    & 71.8   \\
    Marigold~\cite{1}       & Latent diffusion v2
  & 3.22G &  \textbf{24.8}   &  \textbf{30.9}     &   \textbf{45.5}    &      \textbf{70.7}     &  82.2   \\
    Adabins~\cite{8}      & ViT-Base & 897M &    25.5   &   31.6    &  43.6    &     70.5    &  \textbf{83.0}    \\
    Omnidata~\cite{6}         & ViT-Base  & 1.36G &   29.5    & 34.7      &  42.9    &        67.3   &   79.7    \\
    HDN~\cite{5}              & ViT-Large  &  1.36G  &   25.0   &  31.0    &  45.1      &   68.9     & 80.2     \\
    DPT~\cite{3}              & ViT-Large  &  1.36G &  28.1     &  34.0     &   44.8    &69.6  &  81.8   \\
    \midrule
    \textbf{Ours}      & Latent diffusion v2 & 4.49G  & \textbf{23.4}      &  \textbf{29.7}   &  \textbf{46.6}  & \textbf{70.9} & \textbf{84.9}\\
    \bottomrule
  \end{tabular}
\end{table}

As shown in Table~\ref{tab:method_comparison}, we can observe the following:
(1) Methods based on GAN (such as pix2pix) are significantly inferior to other models in MAE and RMSE, indicating that the adaptability of the GAN architecture to the continuous depth value regression task is insufficient, and it is difficult to control the global geometric consistency.
(2) The latent diffusion model (such as Marigold) generates a detailed depth map through iterative denoising, and its $\delta^{3}$ index verifies the advantage of the diffusion model in detail preservation. However, it underperforms our method across other key metrics, but lacks high-level semantic constraints. In areas with complex abrupt changes of the terrain (such as large undulating terrain), it is easy to produce abnormal predictions.
(3) The ViT series models (such as HDN and Adabins) have reached a good level in MAE and RMSE due to the global attention mechanism of Transformer, and Adabins has increased $\delta^{3}$ to 83.0\% (the highest value of the ViT model) through the adaptive depth binning strategy, but its MAE is 2.7\% higher than the Marigold model's, indicating that there is still room for improvement in the control of local details by the dynamic binning mechanism.
(4) Our method achieves comprehensive breakthroughs in various indicators through collaborative optimization of text-visual features, especially MAE index is improved by 5.6\% compared to the Marigold model, which verifies the effectiveness of semantic cross-modal guidance for depth inference in complex scenes.Detailed experimental results could be found in the Appendix~\ref{app:appendixB}.
\subsection{Efficiency Benchmarking}
\begin{figure}[!ht]
  \centering
  \begin{subfigure}[b]{0.48\textwidth}
    \includegraphics[width=\textwidth]{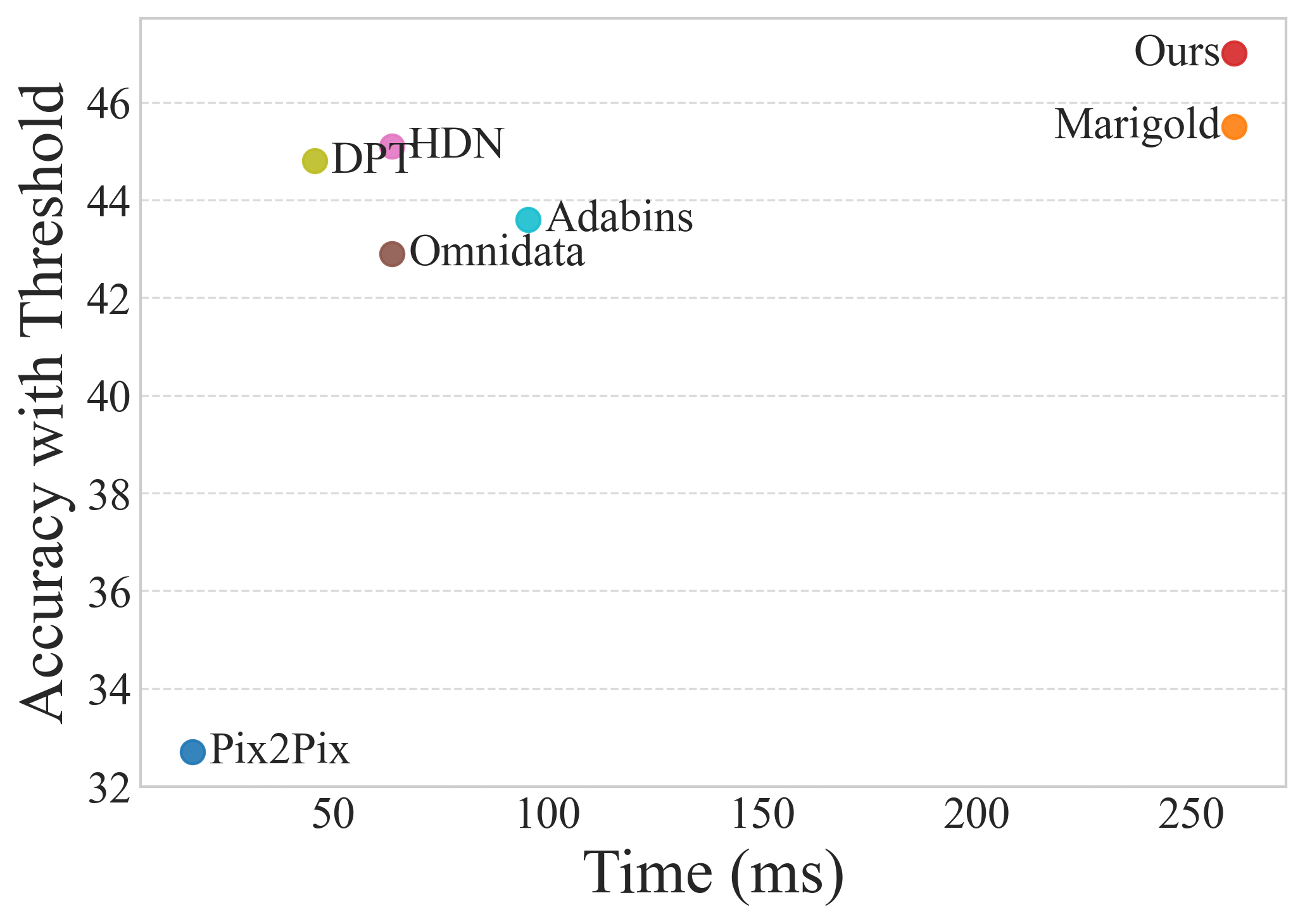}
    \label{fig:bubble}
  \end{subfigure}
  \hspace{0.2em}
  \begin{subfigure}[b]{0.48\textwidth}
    \includegraphics[width=\textwidth]{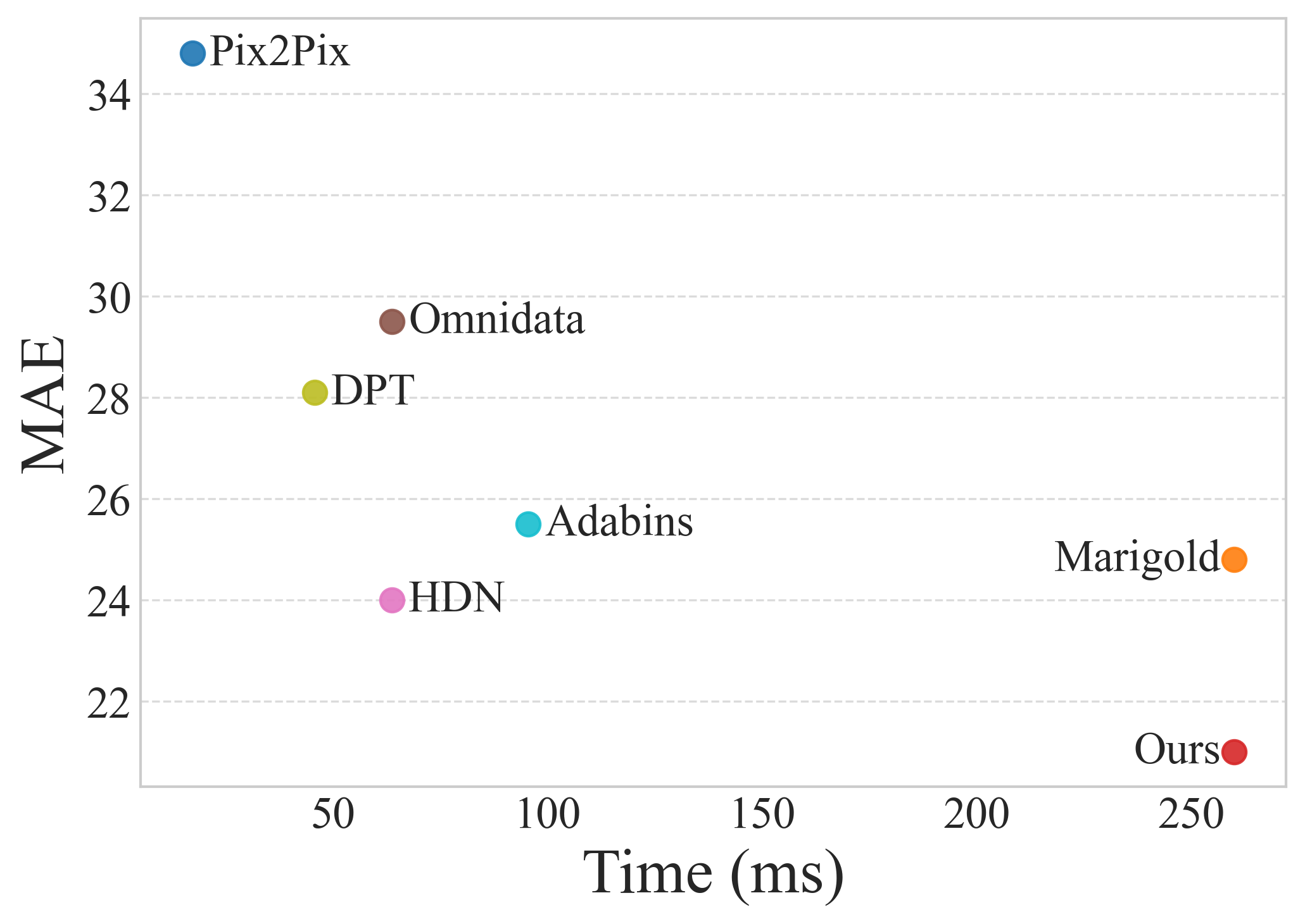}
    \label{fig:radar}
  \end{subfigure}
  \caption{Time consumption and $\delta$ of  depth estimation methods on 30m Resolution Data}
  \label{fig:time}
\end{figure}
After multiple training and evaluations, we have statistically analyzed the relationship between $\delta$ of the depth estimation benchmark model and its time consumption. Figure~\ref{fig:time} indicates that depth estimation method struggle to balance performance with computational efficiency. For instance, Pix2Pix achieves the fastest inference but suffers severe accuracy degradation. Marigold achieves higher accuracy but runs 15 times slower than Pix2Pix during inference. On the other hand, while the HDN method also consumes more time than DPT, it achieves a reasonable trade-off between performance and computational efficiency across both datasets. Detailed experimental results could be found in the Appendix~\ref{app:appendixB}.

\subsection{Model Generalization Across Terrain Variations}
\begin{table}[ht]
  \caption{Cross-terrain generalization:Performance of the model in different terrain areas. (D1: Plain, D2: Mountains)}
  \label{tab:Terrain_Variations_comparison}
  \centering
  \begin{tabular}{lcccccccccc} 
  \toprule
  \textbf{Model} 
    & \multicolumn{2}{c}{\textbf{MAE} $\downarrow$} 
    & \multicolumn{2}{c}{\textbf{RMSE} $\downarrow$} 
    & \multicolumn{2}{c}{\boldmath$\delta$ $\uparrow$}
    & \multicolumn{2}{c}{\boldmath$\delta^{2}$ $\uparrow$}
    & \multicolumn{2}{c}{\boldmath$\delta^{3}$ $\uparrow$}\\
  \cmidrule(lr){2-3} \cmidrule(lr){4-5} \cmidrule(lr){6-7}
  \cmidrule(lr){8-9} \cmidrule(lr){10-11}
    & D1 & D2 & D1 & D2 & D1 & D2 & D1 & D2 & D1 & D2\\
  \midrule
  pix2pix       & 22.6   & 34.5 &   28.8  & 41.4    & 34.7    & 26.1 & 57.7 & 47.1 & 70.3 & 62.7 \\
  Marigold & \textbf{13.9}   & \textbf{24.2} &   \textbf{17.5}  & \textbf{30.4}       & \textbf{54.7}    &  \textbf{38.0} & \textbf{76.5} & \textbf{62.4} & 85.4 & \textbf{76.7} \\
  Adabins & 16.7   & 24.3 & 20.8  & 31.8   & 49.7   & 31.6& 76.1 &57.9 & \textbf{86.5} &73.5 \\
  Omnidata & 16.6  & 30.7 & 21.0  & 36.0       & 48.6    &  33.0 & 75.1 & 54.2 & 81.6 & 67.6 \\
  HDN & 14.0  & 24.2 & 17.5  & 31.4      &53.7    & 33.4 & 73.0 & 55.2 & 80.2 & 68.8\\
  DPT & 16.7  & 34.2 & 21.5  & 42.2       & 48.0    &  27.8 &  71.1 & 48.5 & 82.4 & 62.4 \\
  \midrule
    \textbf{Ours}        & \textbf{12.7}  & \textbf{20.5} & \textbf{16.8}  & \textbf{28.0}      & \textbf{59.5}    & \textbf{42.3} &  \textbf{77.9} & \textbf{66.7} & \textbf{87.5} & \textbf{79.0} \\
  \bottomrule
 \end{tabular}
\end{table}
We divided the terrain into two subsets based on terrain types: one subset is primarily mountains terrain, while the other is primarily plains. As shown in Table~\ref{tab:Terrain_Variations_comparison}, It can be observed that in the plain dataset (D1), where altitude variations are limited, models establish reliable correlations between visual cues and depth values. However, this stability collapses in mountain areas (D2), where the same model has significantly reduced RMSE and $\delta^{3}$, exposing the fragility of purely visual approaches when confronted with abrupt elevation transitions and high-frequency topographic features. For more details on the dataset division, please refer to Appendix~\ref{app:testdata}.  

Our text-conditioned diffusion framework breaks this paradigm by injecting geographic semantics into the depth generation process, The model receives an explicitly injected domain prior knowledge, with the text providing high-level semantic labels (such as \enquote{mountain}), constraining the macro topographical features that the depth map must satisfy.Compared to Marigold, there has been an improvement of 8.6\% and 15.3\% in MAE at D1 and D2, respectively. The text-visual fusion narrows the cross-terrain performance gap. This multi-modal synergy achieves what neither modality accomplishes alone, effectively bridging the accuracy gap between data-driven depth estimation and LiDAR surveys.

\subsection{Text-Driven Depth Estimation}
\begin{table}[ht]
  \caption{Performance comparison of Marigold (with text-form geographic semantics) against other models.}
  \label{tab:text-driven}
  \centering
  \begin{tabular}{l|c|c|c|c|c|c}
    \toprule
    \textbf{Model}       & \textbf{Backbone} &  \textbf{MAE$\downarrow$} &\textbf{RMSE$\downarrow$} & \textbf{\boldmath$\delta$$\uparrow$} & \textbf{\boldmath$\delta^{2}$$\uparrow$} & \textbf{\boldmath$\delta^{3}$$\uparrow$} \\
     \midrule
    pix2pix       & GAN   & 33.5 &   41.2  & 39.8      &   59.6         &  69.1   \\
    Marigold       & Latent diffusion v2 & \textbf{25.0} & \textbf{31.4} & \textbf{44.6} & \textbf{69.2} & \textbf{80.5} \\
    Adabins        & ViT-Base  & 27.0 & 34.2 & 42.1 & 68.3 & 80.0 \\
    Omnidata       & ViT-Base   & 32.3 & 39.0 & 40.5 & 63.2 & 74.5 \\
    HDN            & ViT-Large  & 25.2 & 31.8 & 44.0 & 66.8 & 78.0 \\
    DPT            & ViT-Large  & 30.5 & 37.5 & 43.1 & 66.3 & 78.0 \\
    \midrule
    \textbf{Ours}          & Latent diffusion v2  & \textbf{21.3} & \textbf{29.5} & \textbf{46.2} & \textbf{71.5} & \textbf{85.5} \\
    \bottomrule
  \end{tabular}
\end{table}
To verify the necessity of text-form geographic semantic, We constructed a balanced subset of terrain scene categories for training and evaluation. As shown in Table~\ref{tab:text-driven}, we compared our method with six baselines, and through the integration of textual information, our approach outperformed existing technologies in most cases. Moreover, the model demonstrated superior performance not only in mountains and plain terrains but also exhibited excellent generalization to different terrain scenarios, showcasing outstanding cross-terrain generalization capability. This experiment statistically verifies the effectiveness of geography-based textual semantic guidance.For more details, please refer to Appendix~\ref{app:testdata}.
\section{Conclusion}\label{sec:Conc}
In order to solve the limitation of the lack of spatial information in the field of remote sensing datasets, in this work, we propose RS3DBench dataset, the first remote sensing dataset with the combination of spatial information alignment and remote sensing images and provides rich text annotations for each pair. Based on this dataset, we construct a comprehensive benchmark to evaluate the current state-of-the-art depth estimation models, reveal their advantages and limitations when processing remote sensing images of natural scenes, and demonstrate the practical utility of our dataset in advancing spatial perception research in the field of remote sensing. The experimental results indicate that deep learning models have significant effectiveness in the task of remote sensing depth estimation, and the introduction of textual forms of geographic semantic information is beneficial for remote sensing depth estimation tasks, but also exposes core challenges such as low-texture areas and occlusion. This study lays a data foundation for future research on the application of 3D geospatial intelligence. The current method does not use time series information, so the dataset lacks continuous observations. Future work can be extended to spatiotemporal joint modeling by introducing time series remote sensing data and supporting longitudinal research on land surface change monitoring, and promote the development of 3D geospatial intelligence.


\clearpage
\bibliographystyle{unsrt}
\bibliography{reference}

\begin{thebibliography}{10}

\bibitem{47}
Michalis~A Savelonas, Christos~N Veinidis, and Theodoros~K Bartsokas.
\newblock Computer vision and pattern recognition for the analysis of 2d/3d remote sensing data in geoscience: A survey.
\newblock {\em Remote Sensing}, 14(23):6017, 2022.

\bibitem{48}
James~B Campbell.
\newblock {\em Mapping the Land: Aerial Imagery for Land Use Information. Resource Publications in Geography.}
\newblock ERIC, 1983.

\bibitem{49}
Xiao-Ling Chen, Hong-Mei Zhao, Ping-Xiang Li, and Zhi-Yong Yin.
\newblock Remote sensing image-based analysis of the relationship between urban heat island and land use/cover changes.
\newblock {\em Remote sensing of environment}, 104(2):133--146, 2006.

\bibitem{50}
Rajendra~P Sishodia, Ram~L Ray, and Sudhir~K Singh.
\newblock Applications of remote sensing in precision agriculture: A review.
\newblock {\em Remote sensing}, 12(19):3136, 2020.

\bibitem{51}
Cees~J Van~Westen.
\newblock Remote sensing and gis for natural hazards assessment and disaster risk management.
\newblock {\em Treatise on geomorphology}, 3(15):259--298, 2013.

\bibitem{53}
Alexander Ku, Peter Anderson, Roma Patel, Eugene Ie, and Jason Baldridge.
\newblock Room-across-room: Multilingual vision-and-language navigation with dense spatiotemporal grounding.
\newblock {\em arXiv preprint arXiv:2010.07954}, 2020.

\bibitem{54}
Jacob Krantz, Erik Wijmans, Arjun Majumdar, Dhruv Batra, and Stefan Lee.
\newblock Beyond the nav-graph: Vision-and-language navigation in continuous environments.
\newblock In {\em Computer Vision--ECCV 2020: 16th European Conference, Glasgow, UK, August 23--28, 2020, Proceedings, Part XXVIII 16}, pages 104--120. Springer, 2020.

\bibitem{52}
Xiaoye Liu.
\newblock Airborne lidar for dem generation: some critical issues.
\newblock {\em Progress in physical geography}, 32(1):31--49, 2008.

\bibitem{1}
Bingxin Ke, Anton Obukhov, Shengyu Huang, Nando Metzger, Rodrigo~Caye Daudt, and Konrad Schindler.
\newblock Repurposing diffusion-based image generators for monocular depth estimation.
\newblock In {\em Proceedings of the IEEE/CVF Conference on Computer Vision and Pattern Recognition}, pages 9492--9502, 2024.

\bibitem{2}
Massimiliano Viola, Kevin Qu, Nando Metzger, Bingxin Ke, Alexander Becker, Konrad Schindler, and Anton Obukhov.
\newblock Marigold-dc: Zero-shot monocular depth completion with guided diffusion.
\newblock {\em arXiv preprint arXiv:2412.13389}, 2024.

\bibitem{3}
Ren{\'e} Ranftl, Alexey Bochkovskiy, and Vladlen Koltun.
\newblock Vision transformers for dense prediction.
\newblock In {\em Proceedings of the IEEE/CVF international conference on computer vision}, pages 12179--12188, 2021.

\bibitem{4}
Emmanouil Panagiotou, Georgios Chochlakis, Lazaros Grammatikopoulos, and Eleni Charou.
\newblock Generating elevation surface from a single rgb remotely sensed image using deep learning.
\newblock {\em Remote Sensing}, 12(12):2002, 2020.

\bibitem{5}
Chi Zhang, Wei Yin, Billzb Wang, Gang Yu, Bin Fu, and Chunhua Shen.
\newblock Hierarchical normalization for robust monocular depth estimation.
\newblock {\em Advances in Neural Information Processing Systems}, 35:14128--14139, 2022.

\bibitem{6}
Ainaz Eftekhar, Alexander Sax, Jitendra Malik, and Amir Zamir.
\newblock Omnidata: A scalable pipeline for making multi-task mid-level vision datasets from 3d scans.
\newblock In {\em Proceedings of the IEEE/CVF International Conference on Computer Vision}, pages 10786--10796, 2021.

\bibitem{7}
Ren{\'e} Ranftl, Katrin Lasinger, David Hafner, Konrad Schindler, and Vladlen Koltun.
\newblock Towards robust monocular depth estimation: Mixing datasets for zero-shot cross-dataset transfer.
\newblock {\em IEEE transactions on pattern analysis and machine intelligence}, 44(3):1623--1637, 2020.

\bibitem{8}
Shariq~Farooq Bhat, Ibraheem Alhashim, and Peter Wonka.
\newblock Adabins: Depth estimation using adaptive bins.
\newblock In {\em Proceedings of the IEEE/CVF conference on computer vision and pattern recognition}, pages 4009--4018, 2021.

\bibitem{21}
Lihe Yang, Bingyi Kang, Zilong Huang, Xiaogang Xu, Jiashi Feng, and Hengshuang Zhao.
\newblock Depth anything: Unleashing the power of large-scale unlabeled data.
\newblock In {\em Proceedings of the IEEE/CVF Conference on Computer Vision and Pattern Recognition}, pages 10371--10381, 2024.

\bibitem{22}
Simon de~Moreau, Mathias Corsia, Hassan Bouchiba, Yasser Almehio, Andrei Bursuc, Hafid El-Idrissi, and Fabien Moutarde.
\newblock Doc-depth: A novel approach for dense depth ground truth generation.
\newblock {\em arXiv preprint arXiv:2502.02144}, 2025.

\bibitem{23}
Kihong Park, Seungryong Kim, and Kwanghoon Sohn.
\newblock High-precision depth estimation with the 3d lidar and stereo fusion.
\newblock In {\em 2018 IEEE International Conference on Robotics and Automation (ICRA)}, pages 2156--2163. IEEE, 2018.

\bibitem{24}
Varun~Ravi Kumar, Stefan Milz, Christian Witt, Martin Simon, Karl Amende, Johannes Petzold, Senthil Yogamani, and Timo Pech.
\newblock Monocular fisheye camera depth estimation using sparse lidar supervision.
\newblock In {\em 2018 21st International Conference on Intelligent Transportation Systems (ITSC)}, pages 2853--2858. IEEE, 2018.

\bibitem{25}
Gwangbin Bae, Ignas Budvytis, and Roberto Cipolla.
\newblock Multi-view depth estimation by fusing single-view depth probability with multi-view geometry.
\newblock In {\em Proceedings of the IEEE/CVF Conference on Computer Vision and Pattern Recognition}, pages 2842--2851, 2022.

\bibitem{26}
Rui Peng, Rongjie Wang, Zhenyu Wang, Yawen Lai, and Ronggang Wang.
\newblock Rethinking depth estimation for multi-view stereo: A unified representation.
\newblock In {\em Proceedings of the IEEE/CVF conference on computer vision and pattern recognition}, pages 8645--8654, 2022.

\bibitem{9}
Nathan Silberman, Derek Hoiem, Pushmeet Kohli, and Rob Fergus.
\newblock Indoor segmentation and support inference from rgbd images.
\newblock In {\em Computer Vision--ECCV 2012: 12th European Conference on Computer Vision, Florence, Italy, October 7-13, 2012, Proceedings, Part V 12}, pages 746--760. Springer, 2012.

\bibitem{10}
Angela Dai, Angel~X Chang, Manolis Savva, Maciej Halber, Thomas Funkhouser, and Matthias Nie{\ss}ner.
\newblock Scannet: Richly-annotated 3d reconstructions of indoor scenes.
\newblock In {\em Proceedings of the IEEE conference on computer vision and pattern recognition}, pages 5828--5839, 2017.

\bibitem{35}
Ot{\'a}vio~AB Penatti, Keiller Nogueira, and Jefersson~A Dos~Santos.
\newblock Do deep features generalize from everyday objects to remote sensing and aerial scenes domains?
\newblock In {\em Proceedings of the IEEE conference on computer vision and pattern recognition workshops}, pages 44--51, 2015.

\bibitem{36}
Curtis~E Woodcock and Alan~H Strahler.
\newblock The factor of scale in remote sensing.
\newblock {\em Remote sensing of Environment}, 21(3):311--332, 1987.

\bibitem{27}
Hamid Laga, Laurent~Valentin Jospin, Farid Boussaid, and Mohammed Bennamoun.
\newblock A survey on deep learning techniques for stereo-based depth estimation.
\newblock {\em IEEE transactions on pattern analysis and machine intelligence}, 44(4):1738--1764, 2020.

\bibitem{11}
Bo~Qu, Xuelong Li, Dacheng Tao, and Xiaoqiang Lu.
\newblock Deep semantic understanding of high resolution remote sensing image.
\newblock In {\em 2016 International conference on computer, information and telecommunication systems (Cits)}, pages 1--5. IEEE, 2016.

\bibitem{15}
Xue Li, Guo Zhang, Hao Cui, Shasha Hou, Shunyao Wang, Xin Li, Yujia Chen, Zhijiang Li, and Li~Zhang.
\newblock Mcanet: A joint semantic segmentation framework of optical and sar images for land use classification.
\newblock {\em International Journal of Applied Earth Observation and Geoinformation}, 106:102638, 2022.

\bibitem{28}
Gregoris Liasis and Stavros Stavrou.
\newblock Satellite images analysis for shadow detection and building height estimation.
\newblock {\em ISPRS Journal of Photogrammetry and Remote Sensing}, 119:437--450, 2016.

\bibitem{34}
Xiang Li, Jian Ding, and Mohamed Elhoseiny.
\newblock Vrsbench: A versatile vision-language benchmark dataset for remote sensing image understanding.
\newblock {\em arXiv preprint arXiv:2406.12384}, 2024.

\bibitem{29}
Qiming Zhou and Yumin Chen.
\newblock Generalization of dem for terrain analysis using a compound method.
\newblock {\em ISPRS journal of photogrammetry and remote sensing}, 66(1):38--45, 2011.

\bibitem{30}
Weijia Li, Lingxuan Meng, Jinwang Wang, Conghui He, Gui-Song Xia, and Dahua Lin.
\newblock 3d building reconstruction from monocular remote sensing images.
\newblock In {\em Proceedings of the IEEE/CVF International Conference on Computer Vision}, pages 12548--12557, 2021.

\bibitem{12}
Xiaoqiang Lu, Binqiang Wang, Xiangtao Zheng, and Xuelong Li.
\newblock Exploring models and data for remote sensing image caption generation.
\newblock {\em IEEE Transactions on Geoscience and Remote Sensing}, 56(4):2183--2195, 2017.

\bibitem{13}
Qimin Cheng, Haiyan Huang, Yuan Xu, Yuzhuo Zhou, Huanying Li, and Zhongyuan Wang.
\newblock Nwpu-captions dataset and mlca-net for remote sensing image captioning.
\newblock {\em IEEE Transactions on Geoscience and Remote Sensing}, 60:1--19, 2022.

\bibitem{14}
Zilun Zhang, Tiancheng Zhao, Yulong Guo, and Jianwei Yin.
\newblock Rs5m and georsclip: A large scale vision-language dataset and a large vision-language model for remote sensing.
\newblock {\em IEEE Transactions on Geoscience and Remote Sensing}, 2024.

\bibitem{16}
{Federal Office of Topography swisstopo}.
\newblock Swissalti3d: High-resolution digital elevation models (2m/0.5m).
\newblock Data set, 2023.
\newblock Version 2023, Accessed: 2024-07-20.

\bibitem{17}
{Japan Aerospace Exploration Agency (JAXA)}.
\newblock Aw3d30: {ALOS} world 3d - 30m mesh.
\newblock Data set, 2023.
\newblock Version 3.2, Accessed: 2024-07-20.

\bibitem{19}
Liqiang Lin, Yilin Liu, Yue Hu, Xingguang Yan, Ke~Xie, and Hui Huang.
\newblock Capturing, reconstructing, and simulating: the urbanscene3d dataset.
\newblock In {\em European Conference on Computer Vision}, pages 93--109. Springer, 2022.

\bibitem{18}
Yixuan Li, Lihan Jiang, Linning Xu, Yuanbo Xiangli, Zhenzhi Wang, Dahua Lin, and Bo~Dai.
\newblock Matrixcity: A large-scale city dataset for city-scale neural rendering and beyond.
\newblock In {\em Proceedings of the IEEE/CVF International Conference on Computer Vision}, pages 3205--3215, 2023.

\bibitem{20}
Andreas Geiger, Philip Lenz, and Raquel Urtasun.
\newblock Are we ready for autonomous driving? the kitti vision benchmark suite.
\newblock In {\em 2012 IEEE conference on computer vision and pattern recognition}, pages 3354--3361. IEEE, 2012.

\bibitem{43}
L~Grammatikopoulos, K~Adam, E~Petsa, and G~Karras.
\newblock Camera calibration using multiple unordered coplanar chessboards.
\newblock {\em The International Archives of the Photogrammetry, Remote Sensing and Spatial Information Sciences}, 42:59--66, 2019.

\bibitem{44}
Jo{\~a}o~P Leit{\~a}o, Matthew Moy~de Vitry, Andreas Scheidegger, and J{\"o}rg Rieckermann.
\newblock Assessing the quality of digital elevation models obtained from mini unmanned aerial vehicles for overland flow modelling in urban areas.
\newblock {\em Hydrology and Earth System Sciences}, 20(4):1637--1653, 2016.

\bibitem{45}
Juan~J Ruiz, Luis Diaz-Mas, Francisco Perez, and Antidio Viguria.
\newblock Evaluating the accuracy of dem generation algorithms from uav imagery.
\newblock {\em The International Archives of the Photogrammetry, Remote Sensing and Spatial Information Sciences}, 40:333--337, 2013.

\bibitem{46}
Fabio Baselice, Giampaolo Ferraioli, Vito Pascazio, and Gilda Schirinzi.
\newblock Contextual information-based multichannel synthetic aperture radar interferometry: Addressing dem reconstruction using contextual information.
\newblock {\em IEEE Signal Processing Magazine}, 31(4):59--68, 2014.

\bibitem{37}
David Eigen, Christian Puhrsch, and Rob Fergus.
\newblock Depth map prediction from a single image using a multi-scale deep network.
\newblock {\em Advances in neural information processing systems}, 27, 2014.

\bibitem{38}
Jin~Han Lee, Myung-Kyu Han, Dong~Wook Ko, and Il~Hong Suh.
\newblock From big to small: Multi-scale local planar guidance for monocular depth estimation.
\newblock {\em arXiv preprint arXiv:1907.10326}, 2019.

\bibitem{40}
Vaishakh Patil, Christos Sakaridis, Alexander Liniger, and Luc Van~Gool.
\newblock P3depth: Monocular depth estimation with a piecewise planarity prior.
\newblock In {\em Proceedings of the IEEE/CVF Conference on Computer Vision and Pattern Recognition}, pages 1610--1621, 2022.

\bibitem{39}
Weihao Yuan, Xiaodong Gu, Zuozhuo Dai, Siyu Zhu, and Ping Tan.
\newblock New crfs: neural window fully-connected crfs for monocular depth estimation. arxiv e-prints.
\newblock {\em arXiv preprint arXiv:2203.01502}, 2022.

\bibitem{41}
Jia Ning, Chen Li, Zheng Zhang, Chunyu Wang, Zigang Geng, Qi~Dai, Kun He, and Han Hu.
\newblock All in tokens: Unifying output space of visual tasks via soft token.
\newblock In {\em Proceedings of the IEEE/CVF International Conference on Computer Vision}, pages 19900--19910, 2023.

\bibitem{42}
Zhenyu Li, Xuyang Wang, Xianming Liu, and Junjun Jiang.
\newblock Binsformer: Revisiting adaptive bins for monocular depth estimation.
\newblock {\em IEEE Transactions on Image Processing}, 2024.

\bibitem{31}
Robin Rombach, Andreas Blattmann, Dominik Lorenz, Patrick Esser, and Bj{\"o}rn Ommer.
\newblock High-resolution image synthesis with latent diffusion models.
\newblock In {\em Proceedings of the IEEE/CVF conference on computer vision and pattern recognition}, pages 10684--10695, 2022.

\bibitem{33}
Geoscience Australia.
\newblock An australian digital elevation model (dem) derived from lidar 5-meter grid, 2015.

\bibitem{32}
Team GLM, Aohan Zeng, Bin Xu, Bowen Wang, Chenhui Zhang, Da~Yin, Diego Rojas, Guanyu Feng, Hanlin Zhao, Hanyu Lai, Hao Yu, Hongning Wang, Jiadai Sun, Jiajie Zhang, Jiale Cheng, Jiayi Gui, Jie Tang, Jing Zhang, Juanzi Li, Lei Zhao, Lindong Wu, Lucen Zhong, Mingdao Liu, Minlie Huang, Peng Zhang, Qinkai Zheng, Rui Lu, Shuaiqi Duan, Shudan Zhang, Shulin Cao, Shuxun Yang, Weng~Lam Tam, Wenyi Zhao, Xiao Liu, Xiao Xia, Xiaohan Zhang, Xiaotao Gu, Xin Lv, Xinghan Liu, Xinyi Liu, Xinyue Yang, Xixuan Song, Xunkai Zhang, Yifan An, Yifan Xu, Yilin Niu, Yuantao Yang, Yueyan Li, Yushi Bai, Yuxiao Dong, Zehan Qi, Zhaoyu Wang, Zhen Yang, Zhengxiao Du, Zhenyu Hou, and Zihan Wang.
\newblock Chatglm: A family of large language models from glm-130b to glm-4 all tools, 2024.

\end{thebibliography}



\clearpage
\appendix
\section{Appendix}

\subsection{Details of Dataset}\label{app:appendixA}
\subsubsection{Depth data}\label{app:Depth}
\begin{figure}[htbp]
  \centering
  \begin{subfigure}[b]{0.8\textwidth}
    \includegraphics[width=\textwidth, height=6cm, keepaspectratio]{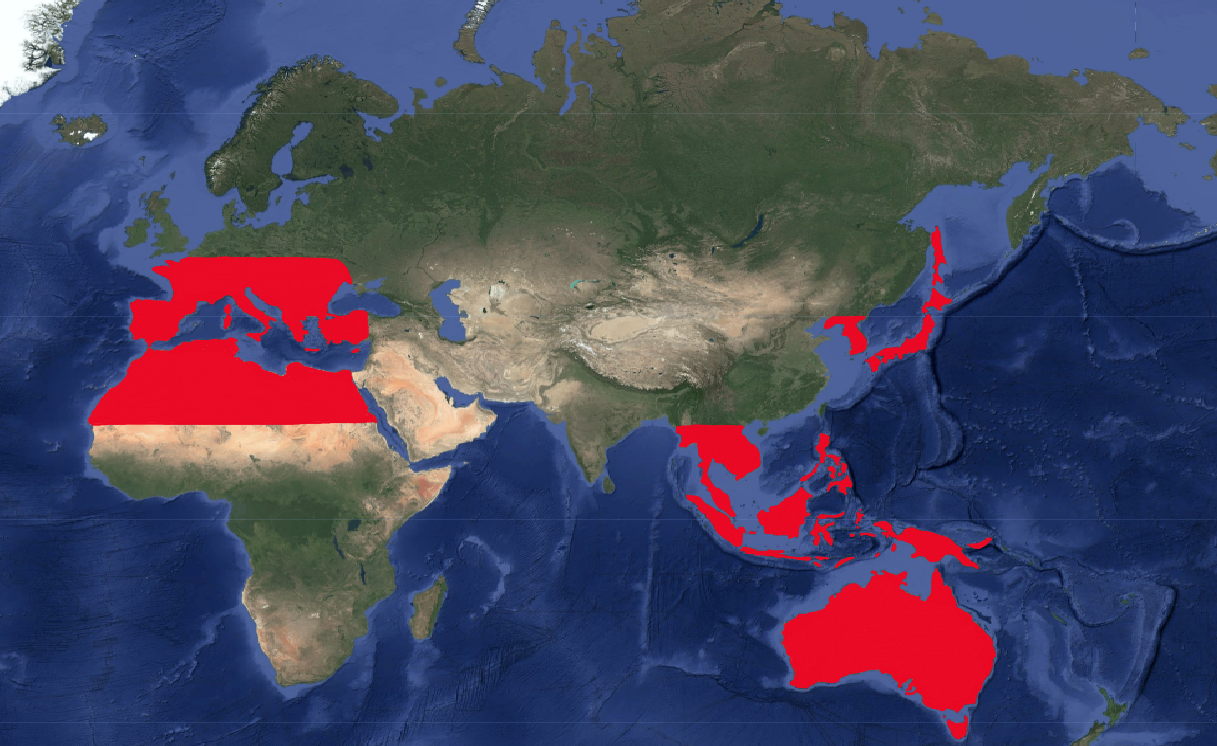}
    \label{fig:geo_dist}
  \end{subfigure}
  
  \caption{Global geospatial coverage}
  \label{fig:geo_coverage}
\end{figure}
Figure~\ref{fig:geo_coverage} shows the geographic coverage, with the dataset spanning four continents, ensuring the adaptability of the model to diverse terrain scenarios. This combination of comprehensive terrain coverage, multi-resolution adaptation, and the authenticity of depth distribution enables the dataset to effectively improve the robustness of remote sensing natural scene depth estimation.The details are as follows:

(1) In Japan, South Korea, Southeast Asia and the Mediterranean, we constructed a DEM dataset based on the advanced land observation satellite ALOS data released by the Japan Aerospace Exploration Agency (JAXA). The data source is the 30 meter spatial resolution global digital elevation model AW3D30, and its data source comes from the three-line sight stereo imaging sensor PRISM carried by the satellite, and its absolute vertical accuracy is RMSE 5 meters (flat area) within the 90\% confidence interval, and the relative vertical accuracy RMSE is 4 meters (undulating terrain). In the data pre-processing, we project and integrate the original data according to the UTM/WGS84 coordinate system, and the data set is stored in the EPSG:4326 coordinate system. The 30m accuracy is very suitable for large-scale terrain and geomorphological analysis, especially in large-scale geographical research, it can provide more detailed terrain undulation information, this data covers the world, we have screened the data of different terrain and landforms from these areas to build different scenes, covering most of the environment required in space research tasks.

(2) In Switzerland, we collected DEMs with an accuracy of 2m and an accuracy of 0.5m from SwissALTI3D, which uses the fusion of airborne LiDAR and aerial photogrammetry to achieve full coverage of a nationwide 2-meter grid resolution (equivalent accuracy ± 0.5 m) under the leadership of the Federal Topographic Service (swisstopo). For alpine areas above 2000 meters above sea level, the point cloud data obtained during the winter snow stabilization period are used to effectively improve the topographic characterization accuracy of glacier fronts and avalanche accumulation areas. Its absolute vertical accuracy is 0.3-0.6 m RMSE in forest cover areas and 0.8-1.2 m RMSE in exposed rock wall areas. The dataset is stored in the Swiss-specific coordinate system LV95 (EPSG: 2056), and the high-precision DEM can provide more ground information, covering the undeveloped areas of Switzerland, and most of them are natural scenes.

(3) In Australia, we systematically collected high-precision geospatial data, with the core dataset derived from the 5-meter spatial resolution DEM and synchronized RGB images from the Google Earth Engine platform. Based on stereo pair data from the Sentinel-2 satellite, the DEM data is generated by photogrammetry and can accurately characterize complex topographic features from coastal lowlands to large watershed mountain ranges. finally built a geographic information database covering part of Australia.
\subsubsection{Data Availability}\label{app:Data}
The Swiss elevation data (2m and 0.5m DEMs) were obtained from the Federal Office of Topography swisstopo (\url{https://www.swisstopo.admin.ch/en/height-model-swissalti3d}). Global terrain data used the 30m AW3D30 DEM provided by the Japan Aerospace Exploration Agency (\url{https://www.eorc.jaxa.jp/ALOS/en/aw3d30}). Australian topography data derives from Geoscience Australia's 5m LiDAR DEM, accessible via DOI: \url{https://doi.org/10.4225/25/5652419862E23}.
\subsubsection{Text annotations}\label{app:text}
We generate text annotations with GLM-v4, and we continuously iterate on generating hints for large language models by generating effects, and we have crafted the following instructions to prompt GLM-v4 to create detailed text annotations: "You are an AI geovision expert who analyzes the types of geography represented by remotely sensed images. Please select several types to describe the remote sensing image: mountains, oceans, lakes, rivers, plains, islands, ridges, farmland. If necessary, provide multiple types." In order to ensure the accuracy of the text annotation of remote sensing images and the rigor of geography, a review team composed of experts in the field of geography and remote sensing was set up in this study.

We divided the terrain of the data into 6 types: Ocean, Plain, Hill, Low undulating mountains, High undulating mountains, and Highland based on the textual description and the height and gradient of the DEM.
\subsubsection{Linear stretching}\label{app:linear}
\begin{figure}[t]
  \centering
  \includegraphics[width=\textwidth]{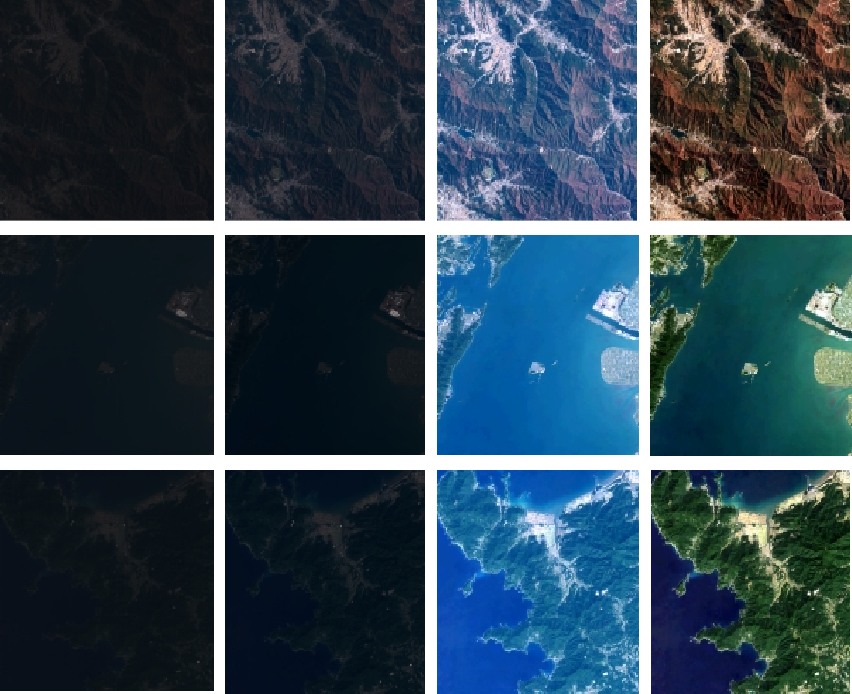}
  \caption{
    \textbf{Comparison of the pre-processing effect of Sentinel-2 images} \\
    (a) direct scaling of raw data (×255); (b) Global normalization to the 0-255 range; (c) Linear stretching enhances contrast;
(d) RGB channels are independently normalized to remove color casts.
  }
  \label{fig:preprocess}
\end{figure}
In order to solve the problem of brightness distribution of Sentinel-2 satellite images, a linear stretching method was used to improve the image contrast. In this method, the effective brightness range is determined by calculating the low percentile (1\%) and high percentile (99\%) of the image, and the pixel values are linearly mapped to the standard range of 0-255 after removing extreme noise. As shown in Figure~\ref{fig:preprocess} : (a) direct scaling of raw data cannot meet the learning needs of the model; (b) Although the global normalization extends the brightness range, the overall color is still dark and has a blue-haze color cast; (c) The contrast is significantly enhanced after linear stretching, and the sharpness of the edges and textures of the ground objects is improved. (d) Further independent normalization of RGB channels to eliminate color casts and restore natural colors, so that the final image is more suitable for model training needs.



The following are the details of the image preprocessing:

(\textbf{i}) Select the effective brightness range: First, calculate the low and high percentiles of the image (such as 1\% and 99\%) to determine the effective brightness range. In this way, the influence of extreme brightness values in the image, such as noise or outliers, on the results can be avoided.

(\textbf{ii})  Cropping: According to the set percentile range, limit the pixel values in the image to the range, and remove the abnormal extreme values.

(\textbf{iii}) Linear mapping: Linear mapping of the pixel value of the image from the current brightness range to the standard gray value range from 0 to 255 by the formula:
\begin{equation}
\mathbf{I_{new}}(x) = \frac{\mathbf{I}(x) - \mathbf{I_{\min}}}{\mathbf{I_{\max}} - \mathbf{I_{\min}}} \times 255,
\end{equation}
where \( \mathbf{I}(x) \) is the original pixel value after cropping, \( \mathbf{I_{\min}} \) and \( \mathbf{I_{\max}} \) are the lowest and highest values after cropping, and \( \mathbf{I_{new}} \) is the stretched pixel value.
where is the original pixel value after cropping, sum is the lowest and highest value after cropping, and is the pixel value after stretching.

Through preprocessing, the brightness range of the image is effectively expanded, thereby enhancing the contrast and detail of the image, and improving the effect of subsequent analysis or model training.

Eventually, we chose to set the elongation to 1\%. The results show that the image quality is significantly improved. However, despite the increased brightness of the image, the image colors still appear dull, with a blue, haze-like hue. The final obtained remote sensing images are closest to human visual perception, so it was selected as the final scheme for RGB channel processing in this study.

The image generated by this processing scheme not only effectively eliminates the color deviation, but also retains the authenticity of the spectral characteristics of the ground objects. It is more in line with the requirements of deep learning models for the distribution of input data. For the color enhancement of multispectral remote sensing images, the channel-independent processing strategy can effectively overcome the color distortion problem caused by the traditional global processing method, and provide a high-quality input data basis for subsequent downstream deep learning tasks.

\subsection{Details of Experiments}\label{app:appendixB}
\subsubsection{Evaluation protocol}
To fully evaluate 3D perception performance against our benchmarks, we employ three common metrics from depth estimation to evaluate depth estimation models:

(1) Threshold Accuracy ($\delta$, $\delta^{2}$, $\delta^{3}$)  
   Measures the percentage of pixels where the predicted depth \( d_i \) and ground truth \( a_i \) satisfy:
\begin{equation}
    \delta^{k} = \frac{1}{M}\sum_{i=1}^M \mathbb{I}\left(\max\left(\frac{d_i}{a_i}, \frac{a_i}{d_i}\right) < 1.25^k\right).
\end{equation}
   where \( k \in \{1,2,3\} \), \( \mathbb{I}(\cdot) \) is the indicator function, and \( M \) denotes valid depth pixels.Higher values indicate better ordinal relationships.

(2) Root Mean Squared Error (RMSE)
   Quantifies relative depth error with squared sensitivity:
\begin{equation}
    \text{RMSE} = \sqrt{\frac{1}{M}\sum_{i=1}^M (d_i - a_i)^2}.
\end{equation}

(3) Mean Absolute Error (MAE)  
   Provides robust measurement of linear depth errors:
\begin{equation}
   \text{MAE} = \frac{1}{M}\sum_{i=1}^M |d_i - a_i|.
\end{equation}
\subsubsection{Test Datasets}\label{app:testdata}
We evaluated 6 benchmark models and our own method on the 3DBench dataset. All experiments were conducted on A6000, with each experiment running for a duration ranging from a few hours to two weeks. The detailed training set division is as follows:
\begin{itemize}[leftmargin=*]
\item[$\bullet$]
To validate the dataset's effectiveness for downstream depth estimation tasks and evaluate the accuracy-efficiency tradeoff of benchmark models, we conducted comprehensive training on 38,875 data pairs with 30-meter spatial resolution. The datasets were divided into training, validation, and test sets, corresponding to 80\%, 10\%, and 10\% of the entire dataset.
\item[$\bullet$]
To validate the performance differences of depth estimation baseline models across varying terrains, we defined two subsets: D1 (predominantly plain terrain with low elevation data and small variance) and D2 (mountain-dominated terrain with high elevation data and large variance). Each subset contains 2,400 pairs. The datasets were divided into training, validation, and test sets, corresponding to 80\%, 10\%, and 10\% of the entire dataset. The criteria for terrain classification are detailed in Appendix ~\ref{app:text}.
\item[$\bullet$]
To validate the effectiveness of text incorporation, we created a subset: a dataset comprising six terrain types, with each type containing 400 samples. We employed a stratified random sampling strategy to partition the dataset into training and test sets: 200 samples (uniformly distributed across each category) were randomly selected from the total samples as the test and validation sets, while the remaining samples were used for model training.
\end{itemize}
\subsubsection{Implementation Details}\label{app:implementation}
\begin{table}[htbp]
\centering
\caption{Model Hyperparameter Configurations}
\label{tab:table3}
\begin{tabular}{@{}llllll@{}}
\toprule
\textbf{Method}    & \textbf{Learning Rate} & \textbf{Batch Size} & \textbf{Optimizer}  & \textbf{DLR} \\ 
\midrule
Marigold      & $3.0 \times 10^{-5}$    & 24            & Adam     & Yes               \\
Marigold-RS  & $3.0 \times 10^{-5}$    & 28            & Adam     & Yes              \\
Omnidata     & $1.0 \times 10^{-5}$    & 10                 & Adam      & No             \\
HDN          & $1.0 \times 10^{-5}$    & 10                 & Adam     & No               \\
DPT          & $3.57 \times 10^{-4}$   & 8                  & AdamW     & No         \\
AdaBins      & $3.57 \times 10^{-4}$   & 8                  & AdamW     & No \\
Pix2Pix      & $2 \times 10^{5}$       & 16                 & Adam      & No \\
\bottomrule
\end{tabular}
\end{table}
Overall Setup: As shown in Table~\ref{tab:table3}, multiple rounds of training were conducted for each model using different hyperparameters. During the training process, the performance of the validation set is monitored in real time after the end of each epoch, and the checkpoint with the best verification performance is finally selected for evaluation. We performed a systematic hyperparameter search for learning rate, batch size, and weight decay. The learning rate is selected from the set $\{1 \times10^{-1}, 1 \times10^{-2}, 1 \times10^{-3}, 3.75 \times 10^{-4}, 5 \times 10^{-5}, 3 \times 10^{-5}, 2 \times 10^{-5}, 1 \times10^{-5}\}$, the batch size is selected from the set$\{2, 4, 8, 10, 14, 16, 24, 28, 32, 64\}$. In addition, the number of epochs is dynamically adjusted based on the actual convergence speed of the model. For the hyperparameters in pix2pix model training, we select from the set to ensure the best experimental results.

Our method: We used Marigold to load the pre-trained stable-diffusion v2 model weights and freeze the VAE, fine-tuning only the U-Net and CLIP encoders. The initial learning rate of the text adapter and U-net is set to $3 \times 10^{-5}$ with the IterExponential scheduler: the first 100 steps are warm-up linearly, then decays exponentially, and drops to 1\% of the initial value at 25 000 steps. The training was performed in parallel on NVIDIA A6000 GPU with a total of 100K iterations in batches of 24, and the input text was first tokenized, truncated, and filled by the CLIPTokenizer to obtain a token ID tensor, which was then encoded into semantic embeddings by the CLIPTextModel. The embedded in the cross-attention modules of each layer of U-Net is combined with the current noise latent variables, and the text conditions are injected into the feature stream through attention weighting, which accurately guides the noise residual prediction, so as to realize the conditional diffusion image generation based on text prompts.

\begin{figure}[htbp]
    \centering
    \includegraphics[width=1\textwidth]{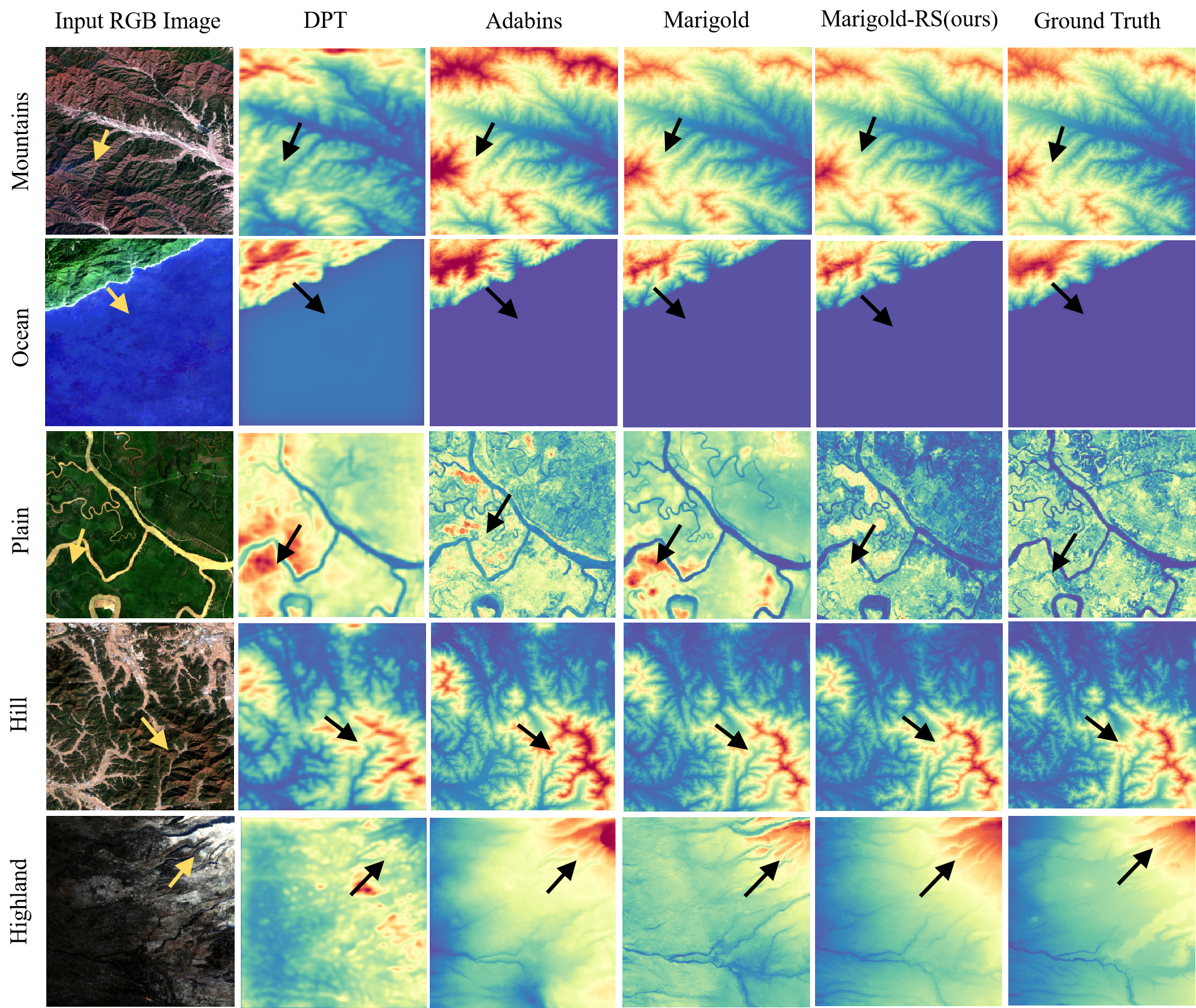}
    \caption{Qualitative comparison of depth estimation methods in different terrain scenarios.}
    \label{fig:output1}
\end{figure}

\begin{figure}[htbp]
    \centering
    \includegraphics[width=1\textwidth]{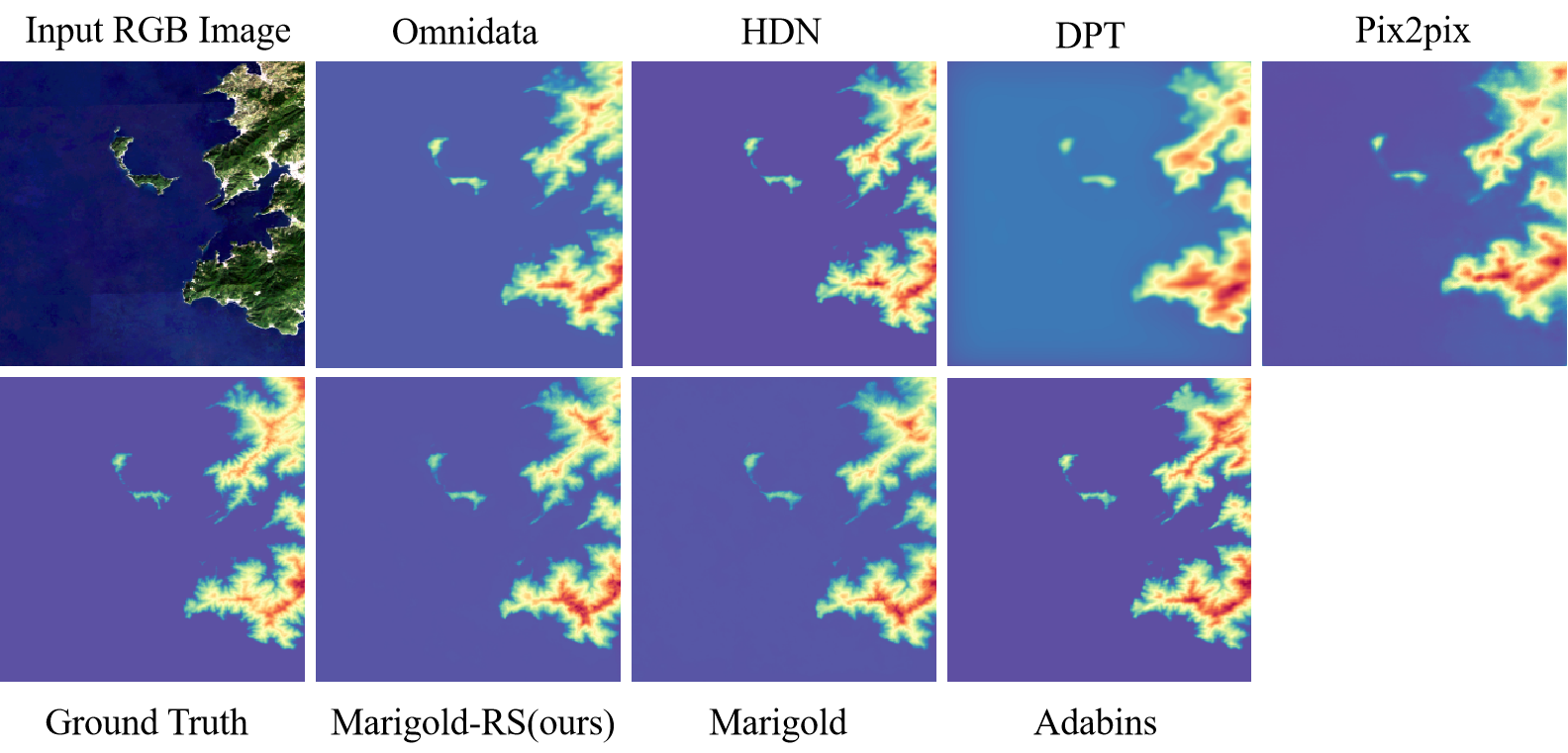}
    \includegraphics[width=1\textwidth]{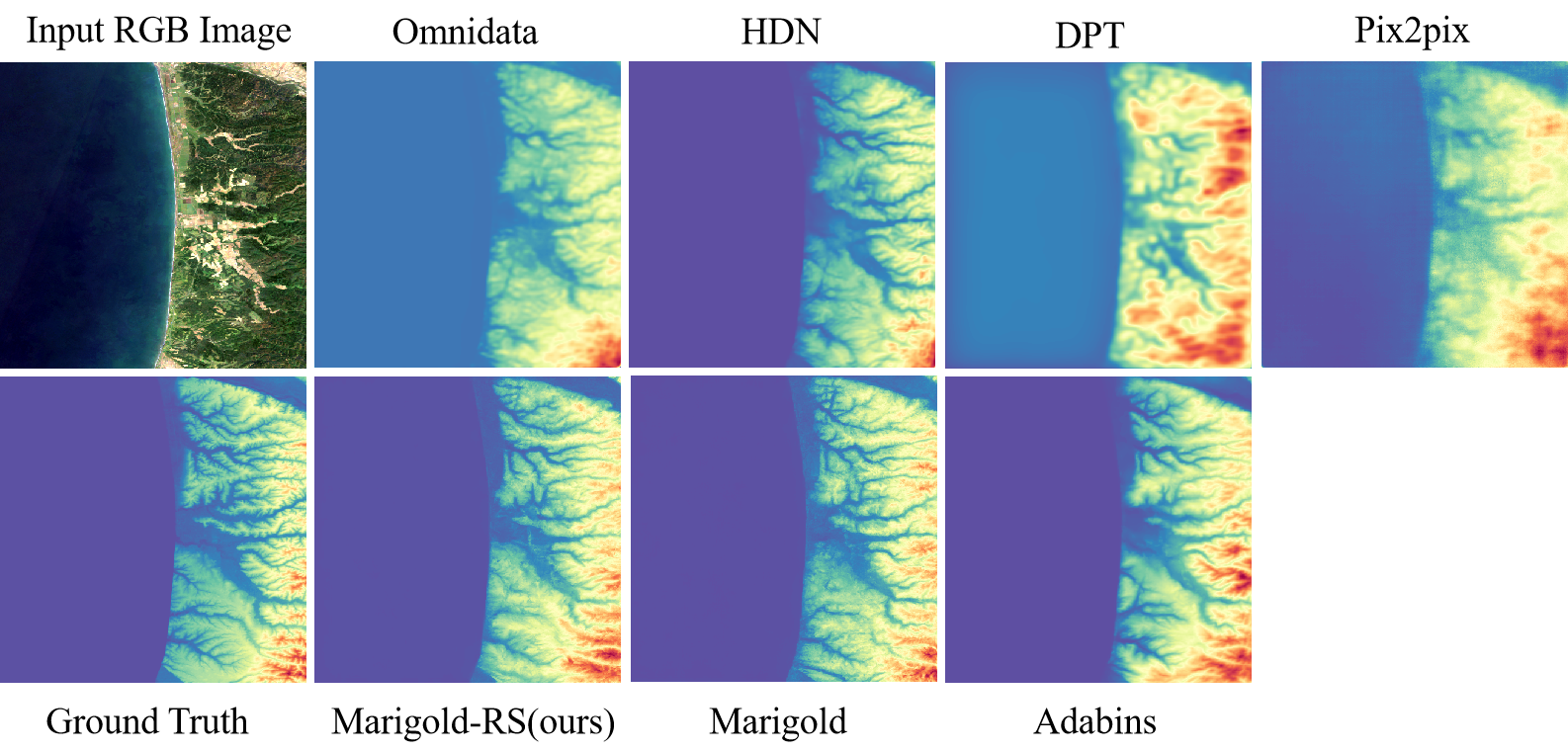}
    \includegraphics[width=1\textwidth]{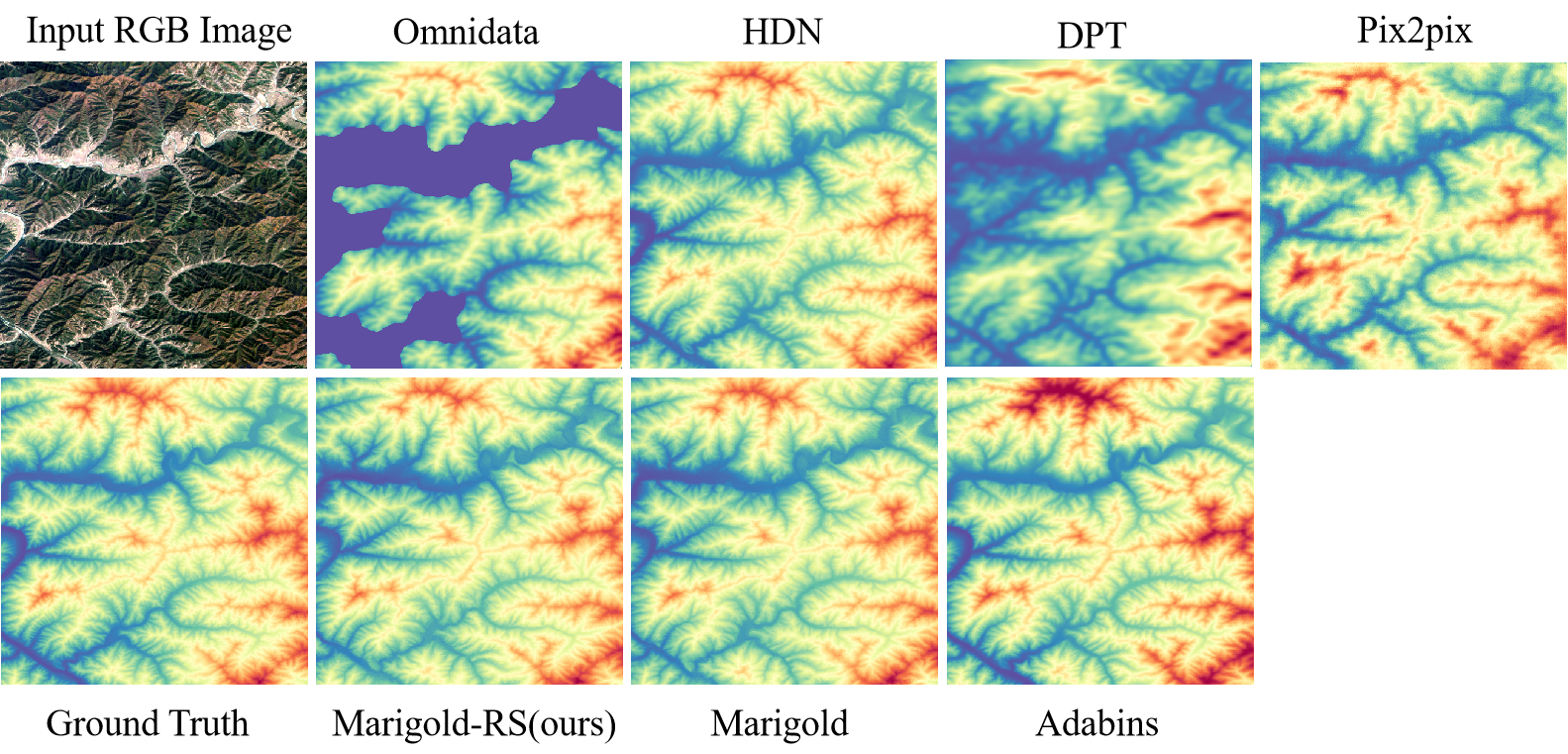}
    \caption{Qualitative comparison of depth estimation methods.}
    \label{fig:output2}
\end{figure}

\clearpage
\newpage
\section*{NeurIPS Paper Checklist}
\begin{enumerate}

\item {\bf Claims}
    \item[] Question: Do the main claims made in the abstract and introduction accurately reflect the paper's contributions and scope?
    \item[] Answer: \answerYes{} 
    \item[] Justification: See Abstract and Section~\ref{sec:intro}.

\item {\bf Limitations}
    \item[] Question: Does the paper discuss the limitations of the work performed by the authors?
    \item[] Answer: \answerYes{} 
    \item[] Justification: See Section~\ref{sec:Conc}.

\item {\bf Theory assumptions and proofs}
    \item[] Question: For each theoretical result, does the paper provide the full set of assumptions and a complete (and correct) proof?
    \item[] Answer: \answerYes{} 
    \item[] Justification: See Section~\ref{sec:Method}.

    \item {\bf Experimental result reproducibility}
    \item[] Question: Does the paper fully disclose all the information needed to reproduce the main experimental results of the paper to the extent that it affects the main claims and/or conclusions of the paper (regardless of whether the code and data are provided or not)?
    \item[] Answer: \answerYes{} 
    \item[] Justification: See GitHub page \url{https://rs3dbench.github.io}.

\item {\bf Open access to data and code}
    \item[] Question: Does the paper provide open access to the data and code, with sufficient instructions to faithfully reproduce the main experimental results, as described in supplemental material?
    \item[] Answer: \answerYes{} 
    \item[] Justification: See GitHub page \url{https://rs3dbench.github.io}.

\item {\bf Experimental setting/details}
    \item[] Question: Does the paper specify all the training and test details (e.g., data splits, hyperparameters, how they were chosen, type of optimizer, etc.) necessary to understand the results?
    \item[] Answer: \answerYes{} 
    \item[] Justification: See Section~\ref{sec:Exper} and Appendix~\ref{app:appendixB}.

\item {\bf Experiment statistical significance}
    \item[] Question: Does the paper report error bars suitably and correctly defined or other appropriate information about the statistical significance of the experiments?
    \item[] Answer: \answerYes{} 
    \item[] Justification: See Appendix ~\ref{app:appendixB}.

\item {\bf Experiments compute resources}
    \item[] Question: For each experiment, does the paper provide sufficient information on the computer resources (type of compute workers, memory, time of execution) needed to reproduce the experiments?
    \item[] Answer: \answerYes{} 
    \item[] Justification: See Section~\ref{sec:Exper} and Appendix~\ref{app:appendixB}.
    
\item {\bf Code of ethics}
    \item[] Question: Does the research conducted in the paper conform, in every respect, with the NeurIPS Code of Ethics \url{https://neurips.cc/public/EthicsGuidelines}?
    \item[] Answer: \answerYes{} 
    \item[] Justification: The research conducted in this article is in line with the NeurIPS Code of Ethics in all respects.

\item {\bf Broader impacts}
    \item[] Question: Does the paper discuss both potential positive societal impacts and negative societal impacts of the work performed?
    \item[] Answer: \answerYes{} 
    \item[] Justification: See Section~\ref{sec:Broader}.
    
\item {\bf Safeguards}
    \item[] Question: Does the paper describe safeguards that have been put in place for responsible release of data or models that have a high risk for misuse (e.g., pretrained language models, image generators, or scraped datasets)?
    \item[] Answer: \answerNA{} 
    \item[] Justification: \answerNA{}

\item {\bf Licenses for existing assets}
    \item[] Question: Are the creators or original owners of assets (e.g., code, data, models), used in the paper, properly credited and are the license and terms of use explicitly mentioned and properly respected?
    \item[] Answer: \answerYes{} 
    \item[] Justification: We indicate licenses of public datasets in Appendix~\ref{app:Data}.

\item {\bf New assets}
    \item[] Question: Are new assets introduced in the paper well documented and is the documentation provided alongside the assets?
    \item[] Answer: \answerYes{} 
    \item[] Justification: We provide the documentation along with the code. See GitHub page \url{https://rs3dbench.github.io}.

\item {\bf Crowdsourcing and research with human subjects}
    \item[] Question: For crowdsourcing experiments and research with human subjects, does the paper include the full text of instructions given to participants and screenshots, if applicable, as well as details about compensation (if any)? 
    \item[] Answer: \answerNA{} 
    \item[] Justification: \answerNA{}

\item {\bf Institutional review board (IRB) approvals or equivalent for research with human subjects}
    \item[] Question: Does the paper describe potential risks incurred by study participants, whether such risks were disclosed to the subjects, and whether Institutional Review Board (IRB) approvals (or an equivalent approval/review based on the requirements of your country or institution) were obtained?
    \item[] Answer: \answerNA{} 
    \item[] Justification: This work does not involve crowdsourcing experiments nor research with human subjects.

\item {\bf Declaration of LLM usage}
    \item[] Question: Does the paper describe the usage of LLMs if it is an important, original, or non-standard component of the core methods in this research? Note that if the LLM is used only for writing, editing, or formatting purposes and does not impact the core methodology, scientific rigorousness, or originality of the research, declaration is not required.
    \item[] Answer: \answerYes{} 
    \item[] Justification:  See Section~\ref{sec:text}.

\end{enumerate}

\end{document}